\documentclass[10pt,twocolumn,letterpaper]{article}

\usepackage{iccv}
\usepackage{times}
\usepackage{epsfig}
\usepackage{graphicx}
\usepackage{amsmath}
\usepackage{amssymb}
\usepackage{paralist}
\usepackage{multirow}
\usepackage{multicol}
\usepackage{adjustbox}
\usepackage{booktabs}
\usepackage{makecell}
\usepackage{xcolor}
\usepackage{bm}
\usepackage{float}
\usepackage{caption}
\usepackage{latexsym}
\usepackage{subcaption}
\usepackage{arydshln}
\graphicspath{ {./fig/} }

\usepackage[pagebackref=true,breaklinks=true,letterpaper=true,colorlinks,bookmarks=false]{hyperref}

\iccvfinalcopy 

\def\iccvPaperID{1558} 

\ificcvfinal\fi

\newcommand*{\affaddr}[1]{#1} 
\newcommand*{\affmark}[1][*]{\textsuperscript{#1}}

\begin{document}

\title{Text to Image Generation with Semantic-Spatial Aware GAN}
\makeatletter
\newcommand{\printfnsymbol}[1]{%
  \textsuperscript{\@fnsymbol{#1}}%
}
\makeatother

\author{%
Kai Hu\affmark[1,\thanks{Kai Hu contributed to this work when he was doing his master thesis supervised by Wentong Liao in TNT.}], Wentong Liao\affmark[1,\thanks{Equal contribution.}], Michael Ying Yang\affmark[2], Bodo Rosenhahn\affmark[1]\\
\normalsize
\affaddr{\affmark[1]TNT, Leibniz University Hannover}, 
\affaddr{\affmark[2]SUG, University of Twente}\\
\normalsize
\vspace{-1em}
\affmark[1]{\tt \{liao,hu,rosenhan\}@tnt.uni-hannover.de},\affmark[2]{\tt \{michael.yang\}@utwente.nl}\\
}

\maketitle
\pagestyle{empty}
\thispagestyle{empty}

\begin{abstract}
    A text to image generation (T2I) model  aims to generate photo-realistic images which are semantically consistent with the text descriptions. 
    Built upon the recent advances in generative adversarial networks (GANs), existing T2I models have made great progress. However, a close inspection of their generated images reveals two major limitations: (1) The condition batch normalization methods are applied on the whole image feature maps equally, ignoring the local semantics; (2) The text encoder is fixed during training, which should be trained with the image generator jointly to learn better text representations for image generation.
    To address these limitations, we propose a novel framework Semantic-Spatial Aware GAN, which is trained in an end-to-end fashion so that the text encoder can exploit better text information. Concretely, we introduce a novel Semantic-Spatial Aware Convolution Network, which (1) learns semantic-adaptive transformation conditioned on text to effectively fuse text features and image features, and (2) learns a mask map in a weakly-supervised way that depends on the current text-image fusion process in order to guide the transformation spatially.
    Experiments on the challenging  COCO and CUB bird datasets demonstrate the advantage of our method over the recent state-of-the-art approaches, regarding both visual fidelity and alignment with input text description.
\end{abstract}

\vspace{-4mm}
\section{Introduction}
\label{sec:intro}

\begin{figure}[t!]
\centering
\begin{minipage}[b]{0.14\textwidth}
\center{\footnotesize{Input Text}}
\end{minipage}
\hfill
\begin{minipage}[b]{0.14\textwidth}
\center{\footnotesize{GT}} 
\end{minipage}
\hfill
\begin{minipage}[b]{0.14\textwidth}
\center{\footnotesize{Generated Image}}
\end{minipage}
\vfill
\begin{minipage}[c]{0.14\textwidth}
\center{ \footnotesize{This is a gray bird with black wings and white wingbars light yellow sides and yellow eyebrows.}}
\end{minipage}
\hfill
\begin{minipage}{0.16\textwidth}
\includegraphics[width=1\textwidth]{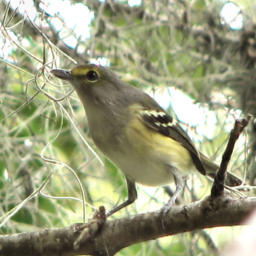}
\end{minipage}
\hfill
\begin{minipage}{0.16\textwidth}
\includegraphics[width=1\textwidth]{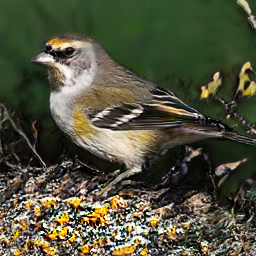}
\end{minipage}
\vfill
\begin{minipage}[c]{0.14\textwidth}
\center{\footnotesize{A horse in a grassy field set against a foggy mountain range.}}
\end{minipage}
\hfill
\begin{minipage}{0.16\textwidth}
\includegraphics[width=\textwidth]{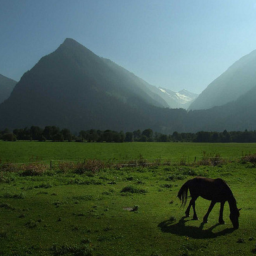}
\end{minipage}
\hfill
\begin{minipage}{0.16\textwidth}
\includegraphics[width=\textwidth]{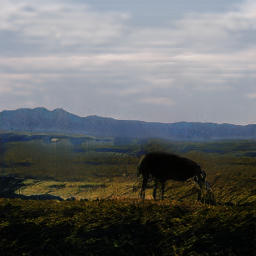}
\end{minipage}
\caption{Examples of images generated by our method (3rd column) conditioned on the given text descriptions.}
\label{fig:teaser}
\vspace{-5mm}
\end{figure}

The great advances made in Generative Adversarial Networks (GANs) \cite{goodfellow2014generative,mirza2014conditional,radford2016unsupervised,zhu2017unpaired,isola2017image,zhang2018stackgan++,brock2019large,karras2019style} boost a remarkable evolution in synthesizing photo-realistic images with diverse conditions, \eg, layout \cite{he2021context}, text \cite{xu2018attngan}, scene graph \cite{ashual2019specifying}, \etc. In particular, generating images conditioned on text descriptions (as shown in Fig. \ref{fig:teaser}) has been catching increasing attention in computer vision and natural language processing communities because: (1) it bridges the gap between these two domains, and (2) linguistic description (text) is the most natural and convenient medium for human being to describe a visual scene. 
Nonetheless, text to image generation (T2I) remains a very challenging task because of the cross-modal problem (text to image transformation) and the ability to keep the generated image semantically consistent with the given text.

The most recent methods for T2I increase the visual quality and resolution by stacking a series of generator-discriminator pairs to generate image from coarse to fine \cite{zhang2017stackgan,zhang2018stackgan++,hong2018inferring,xu2018attngan,li2019control,yin2019semantics}. 
This method has proved effective in synthesizing high-resolution images. However, multiple generator-discriminator pairs lead to higher computation and more unstable training processes. Moreover, the quality of the image generated by the earlier generator decides the final output. If the early generated image is poor, the later generators can not improve its quality. To address this problem, the one-stage generator is introduced in \cite{ming2020DFGAN} which has one generator-discriminator pair. In this work, we also follow this one-stage structure.

Another limitation of current T2I models is to effectively fuse text and image information.
There are three main approaches for the fusion: features concatenation, cross-modal attention
and Condition Batch Normalization (CBN). 
In the early works \cite{reed2016generative,zhang2017stackgan,zhang2018stackgan++}, the text-image fusion is realized by naive concatenation,  which neither sufficiently exploits text information nor effective text-image fusion. 
The most recent works suggest cross-modal attention methods that compute a word-context vector for each sub-region of the image, such as AttnGAN \cite{xu2018attngan}. 
However, the computation cost increases rapidly with larger image size. 
Furthermore, the natural language description is in high-level semantics, while a sub-region of the image is relatively low-level \cite{chen2017sca,yu2017multi}. 
Thus, it cannot explore the high-level semantics well to control the image generation process, especially for complex image with multiple objects. 
The word-level and sentence-level CBNs are proposed in SD-GAN \cite{yin2019semantics} to inject text information in the image feature maps. But their CBNs are only applied a few times during the image generation process so that the text features and image features are not fused sufficiently.
In DF-GAN \cite{ming2020DFGAN}, series of stacked affine transformations, whose parameters are learned from text vector, are used to channel-wise scale and shift the image features.
However, their affine transformations work on the feature maps spatial equally. Ideally, the text information should be only added to the text-relevant sub-regions.
Moreover, all the above methods fix the pre-trained text encoder during training. We argue that this is sup-optimal while if the text encoder could be trained jointly with the image generator, it will better exploit text information for image generation.
Overall, the current text-image fusion methods cannot deeply and efficiently fuse the text information into visual feature maps in order to control the image generation process conditioning on the given text.

\begin{figure*}[t!]
    \centering
    \includegraphics[width=1\textwidth]{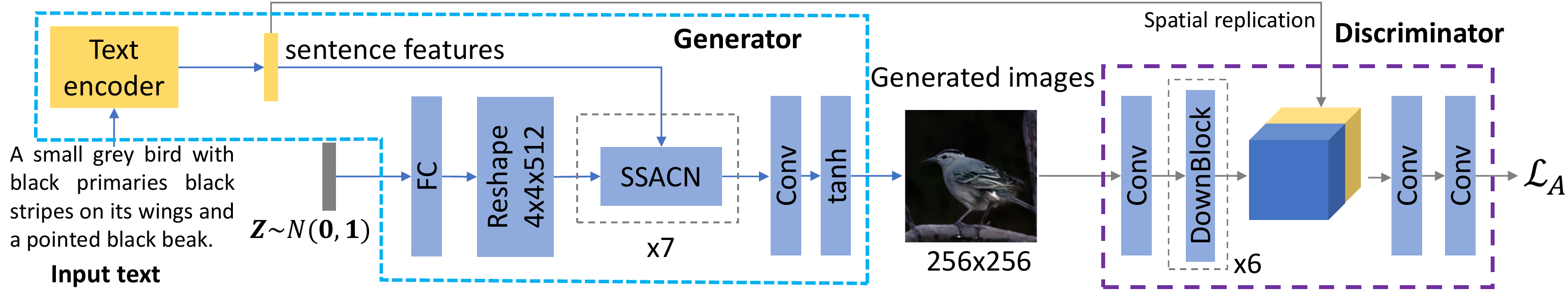}
    \caption{A schematic of our framework SSA-GAN. It has one generator-discriminator pair. The generator mainly consists of 7 proposed SSACN blocks which fuse text and image features through the image generation process and guarantee the semantic text-image consistency. The gray lines indicate the data streams only for training.}
    \label{fig:pipeline}
    \vspace{-2mm}
\end{figure*}

To address the aforementioned issues, we propose a novel T2I framework dubbed as Semantic-Spatial Aware Generative Adversarial
Network (SSA-GAN) (see Fig.~\ref{fig:pipeline}). It has one generator-discriminator pair and is trained in end-to-end fashion so that the pre-trained text encoder is fine tuned to learn better text representations for generating images.
The core element of the framework is the Semantic-Spatial Aware convolution network (SSACN) which consists of a CBN module called Semantic-Spatial Condition Batch Normalization (SSCBN), a residual block, and a mask predictor, as shown in Fig.~\ref{fig:ssac}. 
SSCBN learns semantic-aware affine parameters conditioned on the learned text feature vector. 
The mask map is predicted depending on the current text-image fusion process (\ie output of last SSCBN block). 
This affine transformation fuses the text and image features effectively and deeply, as well as encourages the image features semantically consistent with the text.
The residual block ensures the text-irrelevant parts of the generated image features do not change.
We perform experiments on the challenging benchmarks COCO \cite{lin2014microsoft} and CUB bird dataset \cite{wah2011caltech}  to validate the performance of SSA-GAN for T2I. 
Our SSA-GAN significantly improves the state-of-the-art performance in Inception Score (IS) \cite{salimans2016improved} and Fr\'{e}chet Inception Distance (FID) \cite{heusel2017gans}. 
Extensive ablation studies are conducted to show how the SSCBN works in each step through the image generation process.
In summary, the \textbf{main contributions} of this paper are follows:
\vspace{1mm}
\begin{compactitem}
    \item We propose a novel framework SSA-GAN that can be trained in an end-to-end fashion so that the text encoder is able to learn better text representation for generating better image features. 
    \item A novel SSACN block is introduced to fuse the text and image features effectively and deeply by predicting spatial mask maps to guide the learned text-adaptive affine transformation.
   The SSACN block is trained in a weakly-supervised way, such that no additional annotation is required.
\end{compactitem}

\section{Related Work}
\paragraph{GAN for Text-to-image Generation}
T2I generation is becoming a hot topic both in CV and NLP communities. Generative Adversarial Network (GAN) \cite{goodfellow2014generative} is the most popular model for this task.
Reed \etal \cite{reed2016generative} is the first to use conditional GANs (cGANs) \cite{yu2017unsupervised} to synthesize plausible images from text descriptions.
To improve the resolution of generated images, the StackGAN structure is introduced in \cite{zhang2017stackgan,zhang2018stackgan++}, which stacks multiple generators in sequence in order to generate image from coarse to fine. For training, each generator has its own discriminator for adversarial training. Many recent works follow this structure \cite{xu2018attngan,zhang2018photographic,li2019control} and have made advances.
Zhu \etal \cite{zhu2019dm} applies a dynamic memory module to refine the image quality of the initial stage with multiple iterations.
To overcome the training difficulties in the stacked GANs structure, Ming \etal \cite{ming2020DFGAN} propose a one-stage structure that has only one generator-discriminator pair for T2I generation. Their generator consists of a series of UPBlocks which is specifically designed to upsample the image features in order to generate high-resolution images. Our framework follows this one-stage structure to avoid the problems in the stacked structure.

\vspace{-4mm}
\paragraph{Text-Image Fusion}
In the early T2I works \cite{reed2016generative,zhang2017stackgan,zhang2018stackgan++}, encoded text vector is simply connected to the sampled noise vector or also some intermediate visual feature maps as the input of generators. 
AttnGAN \cite{xu2018attngan} utilizes cross-modal attention to compute a word-context vector for each sub-region of the image and concatenates them to the image feature maps for further text-image fusion. Moreover, it introduces Deep Attentional Multimodal Similarity Model (DAMSM) to measure the image-text similarity both at the word level and sentence level to compute a fine-grained loss for image generation. In this way, the generated image is forced to semantically consistent with the text.
ControlGAN \cite{li2019control} introduces word-level spatial and channel-wise attention block to synthesize sub-regions features corresponding to the most relevant words during the generation process.
DM-GAN \cite{zhu2019dm} utilizes a memory network to dynamically select the important text information based on the current generated image content for further refining the image features. 
Semantic-conditioned batch normalization (BN) is introduced in SD-GAN \cite{yin2019semantics} which conducts a BN conditioned on the global sentence vector and a BN conditioned on the word vectors. 
In DF-GAN \cite{ming2020DFGAN}, on each stage the affine transformation parameters are learned conditioned on the encoded text vector. Then, multiple stacked affine transformations are operated on the image feature maps to fuse the text and image features.
%
In our work, semantic-aware batch normalization is conditioned on text vector (sentence level) which requires much less computation compared to in word level so that it is used through the generation process to deepen the text-image fusion.
The affine transformation is spatially guided by the mask maps predicted from the current image features.
Our idea of SSCBN is inspired by the SPADE in \cite{park2019semantic} which learns pixel-wise batch normalization on feature maps conditioned on input segmentation maps. But our work is essentially different from theirs in two aspects. First, their work is designed for image generation from segmentation maps (pixel to pixel) while ours on text to image. Text is much more abstract and has a larger gap with image compared to segmentation maps. Second, segmentation maps already provide precise spatial information but text does not.

\section{Method}
The architecture of our SSA-GAN is shown in Fig.~\ref{fig:pipeline}. We follow the one-stage structure proposed in \cite{ming2020DFGAN} but replace their UPBlocks with our SSACN blocks.
SSA-GAN has a text encoder that learns text representations, a generator that has $7$ SSACN blocks for deepening text-image fusion process and improving resolution, and a discriminator that is used to judge whether the generated image is semantic consistent to the given text. 
SSA-GAN takes a text description and a noise vector $z\in \mathbb{R}^{100}$ sampled from a normal distribution as input, and outputs a RGB image in size of $256\times256$.
We elaborate each part of our model as follows.

\subsection{Text Encoder}
We adopt the pre-trained text encoder provided by \cite{xu2018attngan} that has been used in many existing works \cite{li2019control,ming2020DFGAN,zhu2019dm}. The text encoder is a bidirectional LSTM \cite{schuster1997bidirectional} and pre-trained using real image-text pairs by minimizing the Deep Attentional Multimodal Similarity Model (DAMSM) loss \cite{xu2018attngan}. It encodes the given text description into a sentence feature vector (the last hidden states of the LSTM) with dimension 256, denoted as $\bar{e}\in \mathbb{R}^{256}$, and word features with length 18 and dimension 256 (the hidden states on each step of the LSTM), denoted as $\mathbf{e}\in \mathbb{R}^{256\times18}$. The $i$-th column $e_i$ of $\mathbf{e}$ is the feature vector of the $i$-th word. In the existing works, this text encoder is adopted by fixing its parameters because it was found that the performance of text to image generation is not improved when the text encoder is fine tuned with the generator \cite{xu2018attngan}. However, we will show in the ablation studies (Sec. \ref{subsec:ablation}) that the text encoder is compatible with our framework for fine tuning so that the performance is further improved.

\subsection{Semantic-Spatial Aware Convolutional Network}
\begin{figure}[t!]
    \centering
    \includegraphics[width=0.5\textwidth]{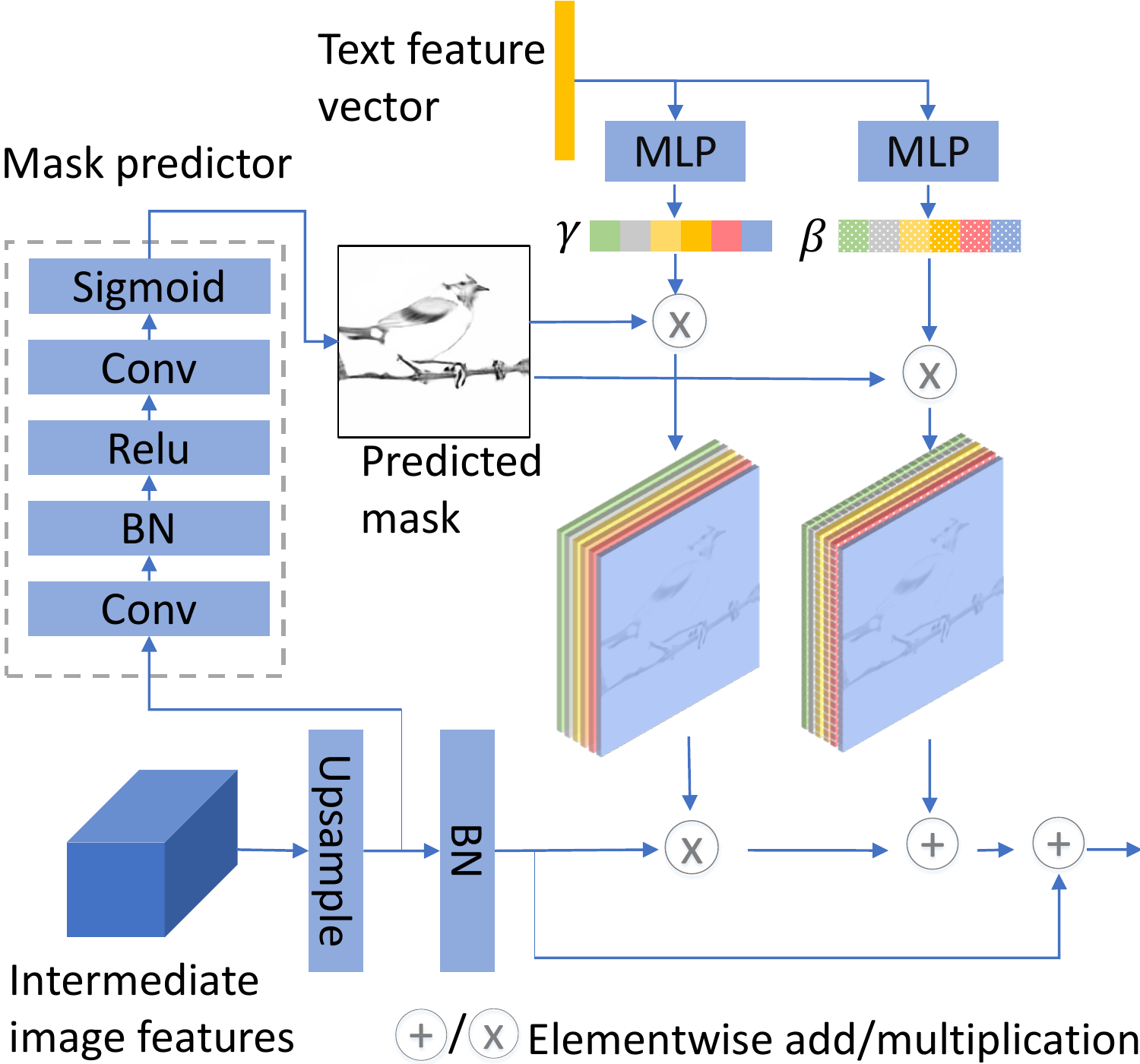}
    \caption{Structure of the SSACN. The SSACN block learns text-aware affine parameters and predicts mask map from current image features in order to realize Semantic-Spatial Condition Batch Normalization.}
    \label{fig:ssac}
    \vspace{-3mm}
\end{figure}

The core of SSA-GAN is the proposed SSACN block as shown in Fig.~\ref{fig:ssac}. It takes the encoded text feature vector $\bar{e}$ and image feature maps $f_{i-1}\in \mathbb{R}^{ch_{i-1} \times \frac{h_i}{2} \times \frac{w_i}{2} }$ from last SSACN block as input, and outputs the image feature maps $f_{i}\in \mathbb{R}^{ch_i\times h_i\times w_i}$ which are further fused with the text features. $w_i$, $h_i$, $ch_i$ are the width, height and number of channels of the image feature maps generated by the $i$-th SSACN block. The input image feature maps of the first SSACN block (no upsampling) are in shape of $4\times4\times512$ which are achieved by projecting the noise vector $z$ to visual domain using a fully-connected (FC) layer and then reshaping it. Therefore, after 6 times upsampling by SSACN blocks, the image feature maps have $256\times256$ resolution. 
Each SSACN block consists of an upsample block, a mask predictor, a Semantic-Spatial Condition Batch Normalization (SSCBN) and a residual block. The upsample block is used to double the width and height of image feature maps by bilinear interpolation operation. The residual block is used to maintain the main contents of the image features to prevent text-irrelevant parts from being changed and the image information is overwhelmed by the text information.
We introduce the Mask Predictor and the SSCBN block in details as follows.

\vspace{-2mm}
\paragraph{Weakly-supervised Mask Predictor}
The structure of the mask predictor is shown in Fig.~\ref{fig:ssac}, as highlighted by the gray dash box. It takes the upsampled image feature maps as input and predicts a mask map $m_i\in\mathbb{R}^{h_i\times w_i}$. The value of any its element $m_{i,(h,w)}$ ranges between $[0,1]$. The value decides how much the following affine transformation should be operated on location $(h,w)$. This map is predicted based on the current generated image feature maps. Thus, it intuitively indicates which parts of the current image feature maps still need to be reinforced with text information so that the refined image feature maps are more semantic consistent to the given text. 
The mask predictor is trained jointly with the whole network without specific loss function to guide its learning process nor additional mask annotation. The only supervision is from the the adversarial loss given by the discriminator which will be discussed in Sec.~\ref{subsec:objective}.
Therefore, it is a weakly-supervised learning process.
In the experiments, we will demonstrate on different stages of SSACN blocks, how the mask map indicates the text-image fusion spatially.

\vspace{-2mm}
\paragraph{Semantic Condition Batch Normalization}
We first give a brief review on standard BN and CBN. Given an input batch $x\in\mathbb{R}^{N\times C\times H\times W}$, where $N$ is the batch size, BN first normalizes it into zero mean and unit deviation for each feature channel:
\vspace{-2mm}
\begin{equation}
\begin{split}
    \hat{x}_{nchw} &= \frac{x_{nchw}-\mu_c(x)}{\sigma_c(x)},\\
    \mu_c(x) &=\frac{1}{NHW}\Sigma_{n,h,w}x_{nchw},\\
    \sigma_c(x) &= \sqrt{\frac{1}{NHW}\Sigma_{n,h,w}(x_{nchw}-\mu_c)^2+\epsilon},
\end{split}
\end{equation}
where $\epsilon$ is a small positive constant for numeric stability. Then, a channel-wise affine transformation is operated:
\begin{equation}
    \tilde{x}_{nchw} = \gamma_c\hat{x}_{nchw}+\beta_c,\label{eq:affine}
\end{equation}
where $\gamma_c$ and $\beta_c$ are learned parameters that work on all spatial locations of all samples in a batch equally. During the test, the learned $\gamma_c$ and $\beta_c$ are fixed.
Apart from using a fixed set of $\gamma$ and $\beta$ learned from training data, Dumoulin \etal \cite{dumoulin2016learned} proposed the CBN which learns the modulation parameters $\gamma$ and $\beta$ adaptive to the given condition for the affine transformation. Then, Eq.~\eqref{eq:affine} can be reformulated as:
\begin{equation}
 \tilde{x}_{nchw} = \gamma(con)\hat{x}_{nchw}+\beta(con).\label{eq:semantic_aware}
\end{equation}

To fuse the text and image features, the modulation parameters $\gamma$ and $\beta$ are learned from the text vector $\bar{e}$:
\begin{equation}
    \gamma_c = P_\gamma(\bar{e}),~~~\beta_c = P_\beta(\bar{e})
\end{equation}
$P_\gamma(\cdot)$ and $P_\beta(\cdot)$ represent the MLPs for $\gamma_c$ and $\beta_c$, respectively. Here, semantic CBN is realized.

\vspace{-2mm}
\paragraph{Semantic-Spatial Condition Batch Normalization}
The semantic aware BN from the last step would work on the image feature maps spatial equally if no more spatial information is added. Ideally, we expect the modulation only works on the text-relevant parts of the feature maps. To realize this, we add the predicted mask map to Eq.~\eqref{eq:semantic_aware} as the spatial condition:
\begin{equation}
 \tilde{x}_{nchw} = m_{i,(h,w)}(\gamma_c(\bar{e})\hat{x}_{nchw}+\beta_c(\bar{e})).\label{eq:spatial_semantic}
\end{equation}
We can see from the formulation that $m_{i,(h,w)}$ does not only decide where to add the text information but also plays the role as weight to decide how much text information needs to be reinforced on the image feature maps.

The modulation parameters $\gamma$ and $\beta$ are learned conditioned on the text information, and the predicted mask maps control the affine transformation spatially. Thus, the Semantic-Spatial CBN is realized in order to fuse the text and image features.

\vspace{2mm}
\subsection{Discriminator}
We adopt the one-way discriminator proposed in \cite{ming2020DFGAN} because of its effectiveness and simplicity. 
The structure of the discriminator is shown in Fig. \ref{fig:pipeline} (in the violet dashed box). It concatenates the features extracted from generated image and the encoded text vector for computing the adversarial loss through two convolution layers. Associated with the Matching-Aware zero-centered Gradient Penalty (MA-GP) \cite{ming2020DFGAN}, it guides our generator to synthesize more realistic images with better text-image semantic consistency. Because the Discriminator is not the contribution of this work, we will not extend its details here and please refer to the paper for more information.

To further improve the quality of generated images and the text-image consistency, and help train the text encoder jointly with the generator, we add the widely applied DAMSM \cite{xu2018attngan} to our framework. Note that, even without the DAMSM, our method already reports the state-of-the-art performance (see Table~\ref{tab:ablation_study_components} in Sec.\ref{sec:exp}). 

\subsection{Objective Functions}
\label{subsec:objective}
\paragraph{Discriminator Objective} 
Since we adopt the one-way discriminator \cite{ming2020DFGAN}, we also use their
adversarial loss associated with the MA-GP loss to train the network. 
\begin{equation}
\begin{split}
\mathcal{L}_{adv}^{D} = &E_{x \backsim p_{data}}[max(0, 1 - D(x, s))] \\
                      &+ \frac{1}{2} E_{x \backsim p_G}[max(0, 1 + D(\hat{x}, s))] \\
                      &+ \frac{1}{2} E_{x \backsim p_{data}}[max(0, 1 + D(x, \hat{s}))]\\
                      &+ \lambda_{MA} E_{x \backsim p_{data}}[(\Arrowvert \triangledown_x D(x, s) \Arrowvert_2 \\
                      &+ \Arrowvert \triangledown_s D(x, s) \Arrowvert_2)^p],
\end{split}
\end{equation}
where $s$ is the given text description while $\hat{s}$ is a mismatched text description. $x$ is the real image corresponding to $s$, and $\hat{x}$ is the generated image. $D(\cdot)$ is the decision given by the discriminator that whether the input image matches the input sentence.
The variables $\lambda_{MA}$ and $p$ are the hyperparameters for MA-GP loss.

\vspace{-4mm}
\paragraph{Generator Objective} The total loss for the generator is composed of an adversarial loss and a DAMSM loss~\cite{xu2018attngan}:
\begin{equation}
\begin{split}
    \mathcal{L}_G &= \mathcal{L}_{adv}^{G} + \lambda_{DA} \mathcal{L}_{DAMSM}\\
    \mathcal{L}_{adv}^{G} &= - E_{x \backsim p_G}[D(\hat{x}, s)],
\end{split}
\end{equation}
where $\mathcal{L}_{DAMSM}$ is a word level fine-grained image-text matching loss~\footnote{Please refer to the Appendix for the detailed expression of $\mathcal{L}_{DAMSM}$~\cite{xu2018attngan}.}, and $\lambda_{DA}$ is the weight of DAMSM loss.



\section{Experiments}
\label{sec:exp}
The effectiveness of our approach is evaluated on the COCO \cite{lin2014microsoft} and CUB bird \cite{wah2011caltech} benchmark datasets, and compared with the recent state-of-the-art GAN methods on T2I generation, StackGAN++ \cite{zhang2018stackgan++}, AttnGAN \cite{xu2018attngan}, ControlGAN \cite{li2019control}, SD-GAN \cite{yin2019semantics}, DM-GAN \cite{zhu2019dm} and DF-GAN \cite{ming2020DFGAN}. Series of ablation studies are conducted to get insight of how each proposed module works.  

\vspace{-2mm}
\paragraph{Datasets}  The CUB bird dataset \cite{wah2011caltech} has 8,855 training images (150  species) and 2,933 test images (50  species). Each bird has 10 text descriptions. 
The COCO dataset \cite{lin2014microsoft} contains 80k training images and 40k test images. Each image has 5 text descriptions. Compared with the CUB dataset,
the images in COCO show complex visual scenes,  making it more challenging for T2I generation tasks.

\vspace{-2mm}
\paragraph{Evaluation Metric}
We adopt the widely used Inception Score (IS) \cite{salimans2016improved} and Fr\'{e}chet Inception Distance (FID) \cite{heusel2017gans} to quantify the performance. 
For the IS scores, a pre-trained Inception v3 network \cite{szegedy2016rethinking} is used to compute the KL-divergence between the conditional class distribution (generated images) and the marginal class distribution (real images). 
A large IS indicates that the generated images are of high quality, and each image clearly belongs to a specific class.
The FID computes the Fr\'{e}chet Distance between the features distribution of the generated and real-world images. 
The features are extracted by a pre-trained Inception v3 network. A lower FID implies the generated images are more realistic. 
To evaluate the IS and FID scores, 30k images in resolution $256\times256$ are generated from each model by randomly selecting text descriptions from the test dataset.
For COCO dataset, previous works \cite{ming2020DFGAN,zhang2020dtgan,li2019object} reported that the IS metric completely fails in evaluating the synthesized images. Therefore, we do not compare the IS on the COCO dataset. The FID is more robust and aligns manually evaluation on the COCO dataset.

\vspace{-2mm}
\paragraph{Implementation details}
Our model is implemented in Pytorch. The batch size is set to 24 distributed on 4 Nvidia RTX 2080-Ti GPUs.
The Adam optimizer \cite{kingma2014adam} with $\beta_1=0.0$ and $\beta_2=0.9$ is used in the training. The learning rates of the generator and the discriminator are set as 0.0001 and 0.0004, respectively. The hyper-parameters $p$, $\lambda_{MA}$ and $\lambda_{DA}$ are set to 6, 2 and 0.1, respectively. 
The model is trained for 600 epoches on CUB dataset and 120 epoches on COCO dataset, respectively.

\begin{table}[t!]
\caption{Performance of IS and FID of StackGAN++, AttnGAN, ControlGAN, SD-GAN, DM-GAN, DF-GAN and our method on the CUB and COCO test set. The results are taken from the authors' own papers.
Best results are in bold.}
\label{tab:results}
\vspace{-2mm}
\centering
\begin{adjustbox}{max width=1\textwidth}
\begin{tabular}{lccc}
\toprule
\multirow{2}*{Methods} &IS $\uparrow$& \multicolumn{2}{c}{FID $\downarrow$} \cr 
    \cmidrule(lr){2-2} \cmidrule(lr){3-4} 
    &CUB&CUB& COCO\\
\midrule 
StackGAN++ \cite{zhang2018stackgan++}   & 4.04 $\pm$ 0.06   &\textbf{15.30}  &81.59\\
AttnGAN \cite{xu2018attngan}            & 4.36 $\pm$ 0.03   &23.98  &35.49\\
ControlGAN \cite{li2019control}         & 4.58 $\pm$ 0.09   & - & -\\
SD-GAN \cite{yin2019semantics}          & 4.67 $\pm$ 0.09   & - & - \\
DM-GAN \cite{zhu2019dm}                 & 4.75 $\pm$ 0.07   &16.09  &32.64\\
DF-GAN \cite{ming2020DFGAN}             & 4.86 $\pm$ 0.04   &19.24 & 28.92\\
Ours& \textbf{5.17 $\pm$ 0.08}& 15.61 & \textbf{19.37}\\
\bottomrule
\end{tabular}
\end{adjustbox}
\vspace{-2mm}
\end{table}

\begin{figure*}[t!]
\centering

\begin{minipage}[c]{0.01\textwidth}
\fontsize{2.0pt}{0.5\baselineskip}\selectfont \center{\ } 
\end{minipage}
\hfill
\begin{minipage}[t]{0.115\textwidth}
\center{\footnotesize{A small bird with an orange bill and grey crown and breast.}}
\end{minipage}
\hfill
\begin{minipage}[t]{0.115\textwidth}
\center{\footnotesize{The bird has a bright red eye, a gray bill and a white neck.}} 
\end{minipage}
\hfill
\begin{minipage}[t]{0.115\textwidth}
\center{\footnotesize{This bird has a long pointed beak with a wide wingspan.}}
\end{minipage}
\hfill
\begin{minipage}[t]{0.115\textwidth}
\center{\footnotesize{A small bird with a black bill and a fuzzy white crown nape throat and breast.}}
\end{minipage}
\vline
\hspace{1pt}
\begin{minipage}[t]{0.115\textwidth}
\center{\footnotesize{A close up of a boat on a field with a cloudy sky.}} 
\end{minipage}
\hfill
\begin{minipage}[t]{0.115\textwidth}
\center{\footnotesize{Some cows are standing on the field on a sunny day.}} 
\end{minipage}
\hfill
\begin{minipage}[t]{0.115\textwidth}
\center{\footnotesize{A skier walks through the snow up the slope.}}
\end{minipage}
\hfill
\begin{minipage}[t]{0.115\textwidth}
\center{\footnotesize{A herd of elephants are walking through a river.}}
\end{minipage}
\vspace{2pt}

\begin{minipage}[c]{0.01\textwidth}
\center{\rotatebox{90}{GT}}
\end{minipage}
\hfill
\begin{minipage}{0.115\textwidth}
\includegraphics[width=\textwidth]{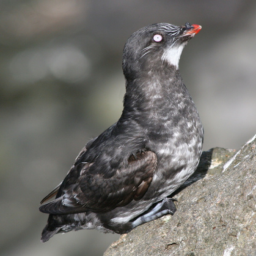}
\end{minipage}
\hfill
\begin{minipage}{0.115\textwidth}
\includegraphics[width=\textwidth]{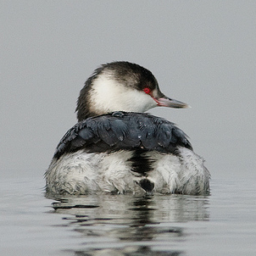}
\end{minipage}
\hfill
\begin{minipage}{0.115\textwidth}
\includegraphics[width=\textwidth]{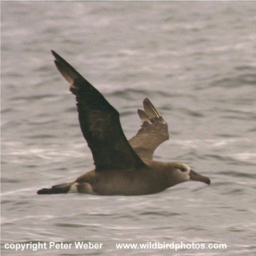}
\end{minipage}
\hfill
\begin{minipage}{0.115\textwidth}
\includegraphics[width=\textwidth]{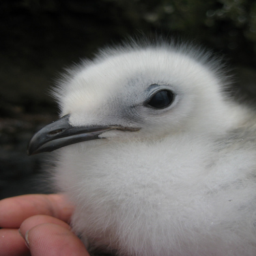}
\end{minipage}
\vline
\hspace{1pt}
\begin{minipage}{0.115\textwidth}
\includegraphics[width=\textwidth]{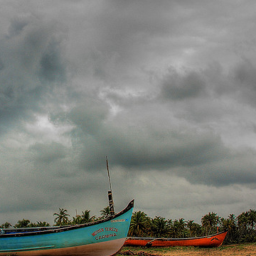}
\end{minipage}
\hfill
\begin{minipage}{0.115\textwidth}
\includegraphics[width=\textwidth]{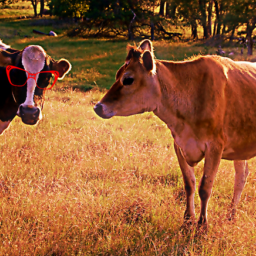}
\end{minipage}
\hfill
\begin{minipage}{0.115\textwidth}
\includegraphics[width=\textwidth]{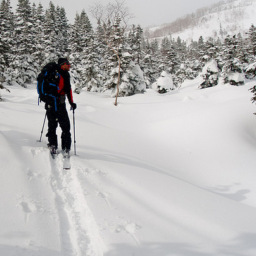}
\end{minipage}
\hfill
\begin{minipage}{0.115\textwidth}
\includegraphics[width=\textwidth]{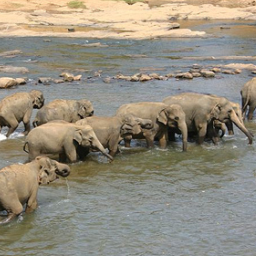}
\end{minipage}
\vspace{5pt}

\begin{minipage}[c]{0.01\textwidth}
\center{\rotatebox{90}{DM-GAN}}
\end{minipage}
\hfill
\begin{minipage}{0.115\textwidth}
\includegraphics[width=\textwidth]{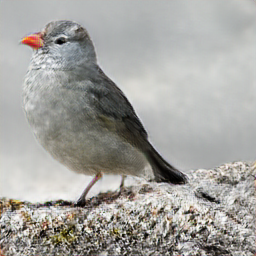}
\end{minipage}
\hfill
\begin{minipage}{0.115\textwidth}
\includegraphics[width=\textwidth]{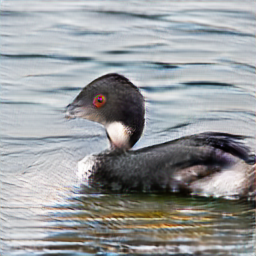}
\end{minipage}
\hfill
\begin{minipage}{0.115\textwidth}
\includegraphics[width=\textwidth]{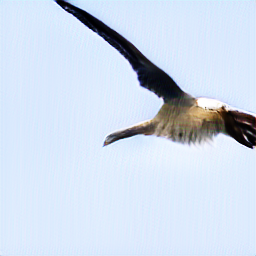}
\end{minipage}
\hfill
\begin{minipage}{0.115\textwidth}
\includegraphics[width=\textwidth]{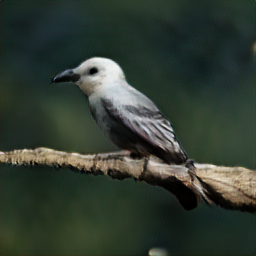}
\end{minipage}
\vline
\hspace{1pt}
\begin{minipage}{0.115\textwidth}
\includegraphics[width=\textwidth]{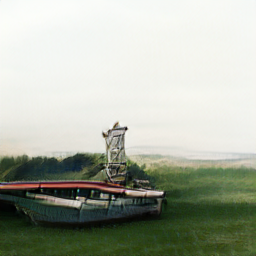}
\end{minipage}
\hfill
\begin{minipage}{0.115\textwidth}
\includegraphics[width=\textwidth]{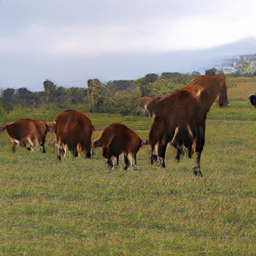}
\end{minipage}
\hfill
\begin{minipage}{0.115\textwidth}
\includegraphics[width=\textwidth]{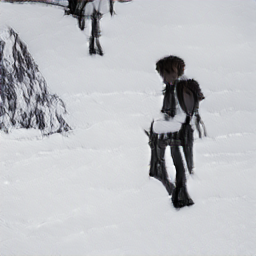}
\end{minipage}
\hfill
\begin{minipage}{0.115\textwidth}
\includegraphics[width=\textwidth]{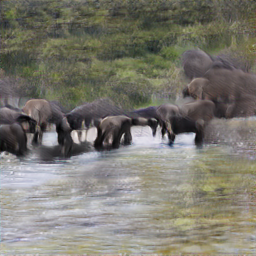}
\end{minipage}
\vspace{2pt}

\begin{minipage}[c]{0.01\textwidth}
\center{\rotatebox{90}{DF-GAN}}
\end{minipage}
\hfill
\begin{minipage}{0.115\textwidth}
\includegraphics[width=\textwidth]{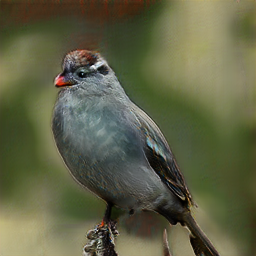}
\end{minipage}
\hfill
\begin{minipage}{0.115\textwidth}
\includegraphics[width=\textwidth]{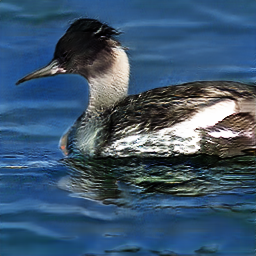}
\end{minipage}
\hfill
\begin{minipage}{0.115\textwidth}
\includegraphics[width=\textwidth]{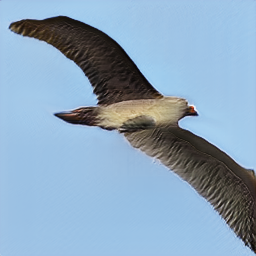}
\end{minipage}
\hfill
\begin{minipage}{0.115\textwidth}
\includegraphics[width=\textwidth]{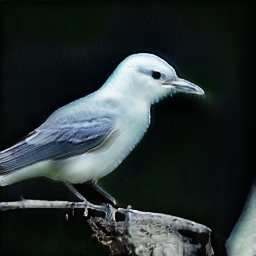}
\end{minipage}
\vline
\hspace{1pt}
\begin{minipage}{0.115\textwidth}
\includegraphics[width=\textwidth]{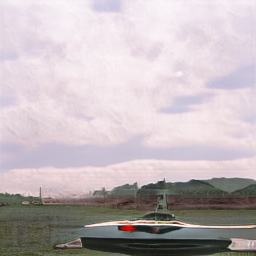}
\end{minipage}
\hfill
\begin{minipage}{0.115\textwidth}
\includegraphics[width=\textwidth]{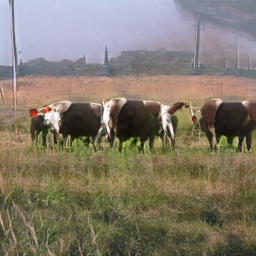}
\end{minipage}
\hfill
\begin{minipage}{0.115\textwidth}
\includegraphics[width=\textwidth]{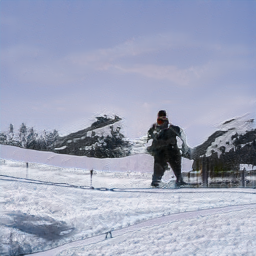}
\end{minipage}
\hfill
\begin{minipage}{0.115\textwidth}
\includegraphics[width=\textwidth]{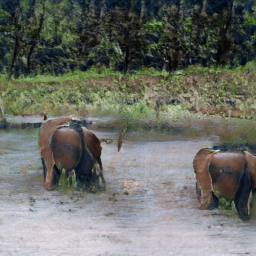}
\end{minipage}
\vspace{2pt}

\begin{minipage}[c]{0.01\textwidth}
\center{\rotatebox{90}{Ours}}
\end{minipage}
\hfill
\begin{minipage}{0.115\textwidth}
\includegraphics[width=\textwidth]{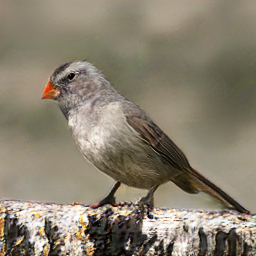}
\end{minipage}
\hfill
\begin{minipage}{0.115\textwidth}
\includegraphics[width=\textwidth]{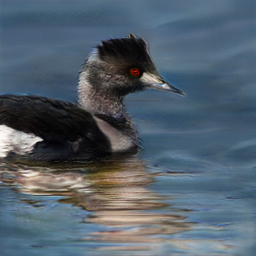}
\end{minipage}
\hfill
\begin{minipage}{0.115\textwidth}
\includegraphics[width=\textwidth]{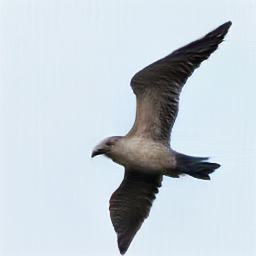}
\end{minipage}
\hfill
\begin{minipage}{0.115\textwidth}
\includegraphics[width=\textwidth]{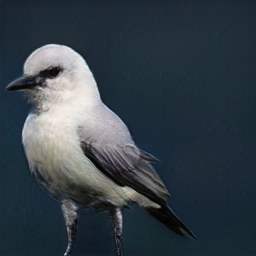}
\end{minipage}
\vline
\hspace{1pt}
\begin{minipage}{0.115\textwidth}
\includegraphics[width=\textwidth]{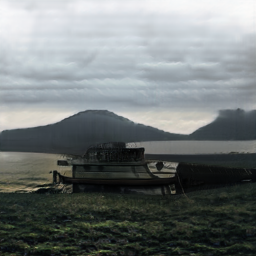}
\end{minipage}
\hfill
\begin{minipage}{0.115\textwidth}
\includegraphics[width=\textwidth]{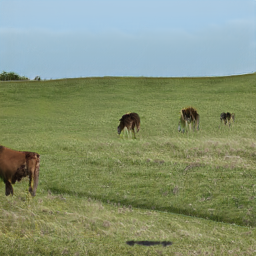}
\end{minipage}
\hfill
\begin{minipage}{0.115\textwidth}
\includegraphics[width=\textwidth]{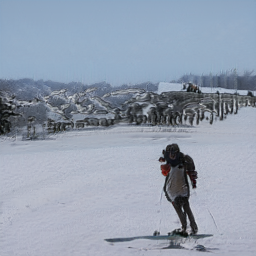}
\end{minipage}
\hfill
\begin{minipage}{0.115\textwidth}
\includegraphics[width=\textwidth]{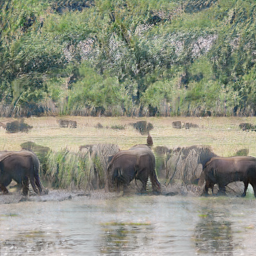}
\end{minipage}
\vspace{2pt}
\caption{Qualitative comparison between our method and the recent state-of-the-art methods DM-GAN \cite{zhu2019dm}, DF-GAN \cite{ming2020DFGAN} on the test set of CUB bird dataset (1st - 4th columns) and COCO dataset (5th - 8th columns). The input text descriptions are given in the first row and the corresponding generated images from different methods are shown in the same column. Best view in color and zoom in.}
\label{fig:qualitative_cub}
\vspace{-4mm}
\end{figure*}

\subsection{Quantitative Results}
Table~\ref{tab:results} shows the quantitative results of our SSA-GAN and several state-of-the-art GAN models that have achieved remarkable advances in T2I generation. 
From the second column of the table we can see that, our SSA-GAN reports the significant improvements in IS (from $4.86$ to $5.17$) on CUB dataset compared to the most recent state-of-the-art method DF-GAN \cite{ming2020DFGAN}. Higher IS means higher quality and text-image semantic consistency. Thus, the superior performance of our method demonstrates that SSA-GAN effectively fuses the text and image features and transforms the text information into images.

Our method remarkably decreases the FID score from $28.92$ to $19.37$  on COCO dataset compared to the state-of-the-art performance.
On CUB dataset, our FID score is a little inferior to the ones given by StackGAN++ \cite{zhang2018stackgan++} ($15.61$ \textit{v.s.} $15.30$) but much lower than the other recent methods: $19.24$ in DF-GAN and $16.09$ in DM-GAN \cite{zhu2019dm}. Compared with the CUB dataset, the COCO dataset is more challenging because there are always multiple objects in images and the background is more complex. Our superior performances indicate that SSA-GAN is able to synthesize complex images in high quality.

The superiority and effectiveness of our proposed SSA-GAN are demonstrated by the extensive quantitative evaluation results that SSA-GAN is able able to generate high-quality images with better semantic consistency, both for the images with many detailed attributes and more complex images with multiple objects.

\begin{figure*}[tb!]
\centering
\begin{minipage}{0.17\textwidth}
\begin{small}
This small bird has a short beak, a light gray breast, a darker gray and black wing tips.
\end{small}
\end{minipage}
\begin{minipage}{0.82\textwidth}
\includegraphics[width=0.12\textwidth]{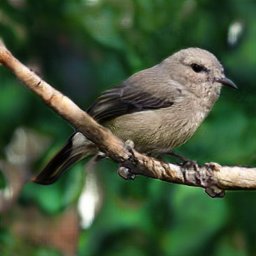}
\includegraphics[width=0.12\textwidth]{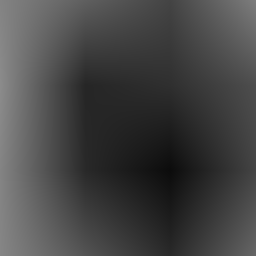}
\includegraphics[width=0.12\textwidth]{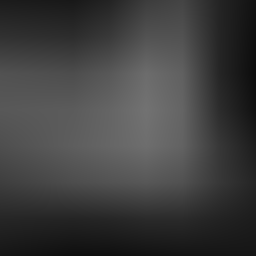}
\includegraphics[width=0.12\textwidth]{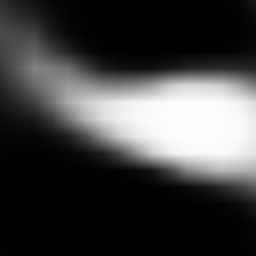}
\includegraphics[width=0.12\textwidth]{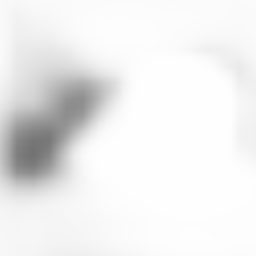}
\includegraphics[width=0.12\textwidth]{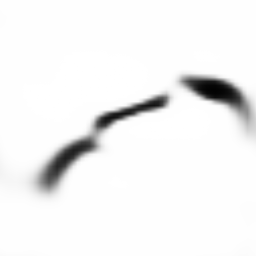}
\includegraphics[width=0.12\textwidth]{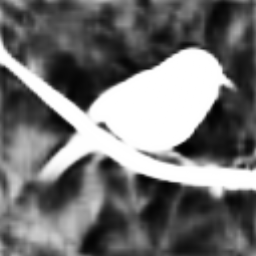}
\includegraphics[width=0.12\textwidth]{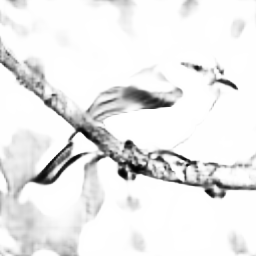}
\end{minipage}
\caption{Example of mask maps predicted in different SSACN blocks. From left to right: input text, generated image and the 7 predicted mask maps (from shallower to deeper layer). Best view in color and zoom in.}
\label{fig:cub_mask_layers}
\vspace{1mm}
\end{figure*}

\begin{figure*}[t!]
\centering
\begin{minipage}{\textwidth}
\center{\textbf{Input text:} A colorful $<$\textbf{color}$>$ bird has wings with dark stripes and small eyes.}
\end{minipage}
\vfill

\begin{minipage}{0.11\textwidth}
\center{\small{\ }}
\end{minipage}
\hfill
\begin{minipage}{0.11\textwidth}
\center{\small{blue}}
\end{minipage}
\hfill
\begin{minipage}{0.11\textwidth}
\center{\small{\ }}
\end{minipage}
\hfill
\begin{minipage}{0.11\textwidth}
\center{\small{red}}
\end{minipage}
\hfill
\begin{minipage}{0.11\textwidth}
\center{\small{\ }}
\end{minipage}
\hfill
\begin{minipage}{0.11\textwidth}
\center{\small{white}}
\end{minipage}
\hfill
\begin{minipage}{0.11\textwidth}
\center{\small{\ }}
\end{minipage}
\hfill
\begin{minipage}{0.11\textwidth}
\center{\small{pink}}
\end{minipage}
\vfill

\begin{minipage}{0.11\textwidth}
\includegraphics[width=\textwidth]{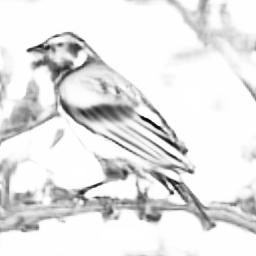}
\end{minipage}
\hfill
\begin{minipage}{0.11\textwidth}
\includegraphics[width=\textwidth]{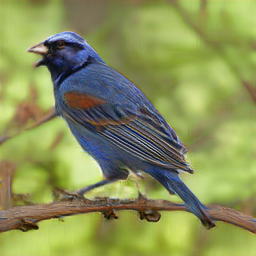}
\end{minipage}
\hfill
\begin{minipage}{0.11\textwidth}
\includegraphics[width=\textwidth]{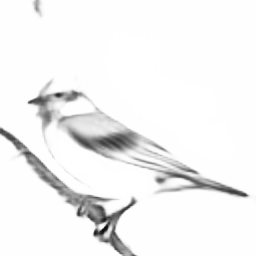}
\end{minipage}
\hfill
\begin{minipage}{0.11\textwidth}
\includegraphics[width=\textwidth]{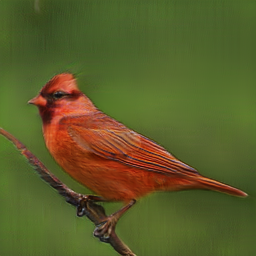}
\end{minipage}
\hfill
\begin{minipage}{0.11\textwidth}
\includegraphics[width=\textwidth]{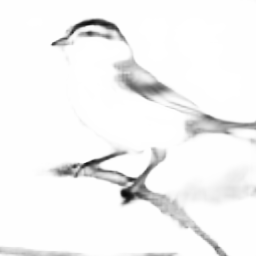}
\end{minipage}
\hfill
\begin{minipage}{0.11\textwidth}
\includegraphics[width=\textwidth]{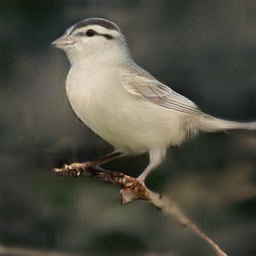}
\end{minipage}
\hfill
\begin{minipage}{0.11\textwidth}
\includegraphics[width=\textwidth]{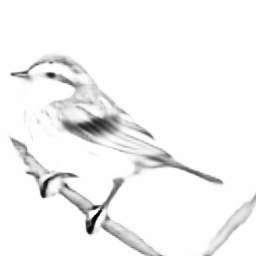}
\end{minipage}
\hfill
\begin{minipage}{0.11\textwidth}
\includegraphics[width=\textwidth]{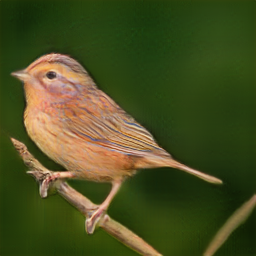}
\end{minipage}
\caption{Examples of diverse image generation by changing the color of the input text  from our method on the test set of CUB dataset. The odd columns show the final predicted masks, and the even columns show the corresponding generated images. Best view in color and zoom in.}
\label{fig:cub_change_color}
\vspace{-3mm}
\end{figure*}

\subsection{Qualitative Results}
We compare the generated images from our method and two state-of-the-art GAN models,~i.e. DM-GAN \cite{zhu2019dm} and DF-GAN \cite{ming2020DFGAN},  for qualitative evaluation, as shown in Fig.~\ref{fig:qualitative_cub}~\footnote{More qualitative examples are given in the Appendix.}.

For the CUB Bird dataset, shown in the first 4 columns in Fig.~\ref{fig:qualitative_cub}, our SSA-GAN generates images with more vivid details that are semantically consistent with the given text descriptions as well as clearer backgrounds. 
For example, in the 1st column, given text ``A small bird with an orange bill and grey crown and breast'', our method generates an image that has all the mentioned attributes. However, the image generated by DM-GAN does not reflect ``small'' while the image generated by DF-GAN does not have ``grey crown and breast''. 
More limitations of other methods can be observed in other examples. DF-GAN can neither generate the ``red eye'' in the 2nd column nor the ``black bill'' in the 4th column.
The birds generated by DM-GAN in the 2nd and 3rd columns are not natural or photo-realistic.
The qualitative results demonstrate that our SSA-GAN is more effectively and deeply to fuse text and image features and has higher text-image consistency. It is good at synthesizing details of a bird described by the text description.

\begin{table}[t!]
\caption{Ablation study of evaluating the impact of SSACN and DAMSM in our framework on the test set of CUB dataset.}
\label{tab:ablation_study_components}
\vspace{-4mm}
\begin{center}
\begin{tabular}{c c c c c}
\hline
\multirow{2}{*}{ID} & \multicolumn{2}{c}{Components} & \multirow{2}{*}{IS $\uparrow$} & \multirow{2}{*}{FID $\downarrow$} \\
\cline{2-3} & SSACN & DAMSM \\
\hline 
0 & - & - & 4.86 $\pm$ 0.04 & 19.24 \\
\hline
1 & \checkmark & - & 4.97 $\pm$ 0.09 & 18.54 \\            
\hline
2 & \checkmark & \checkmark & 5.07 $\pm$ 0.04 & \textbf{15.61} \\
\hline 
3 & \checkmark & \checkmark (fine-tune) & \textbf{5.17 $\pm$ 0.08} & 16.58 \\
\hline
\end{tabular}
\end{center}
\vspace{-4mm}
\end{table}

For the COCO dataset, shown in the last 4 columns in Fig.~\ref{fig:qualitative_cub}, one can observe that SSA-GAN is able to generate complex images with multiple objects with different backgrounds. In the 5th column, our image is more realistic than the ones generated by DM-GAN and DF-GAN. 
In 6th column, each of the generated cows can be clearly recognized and separated, while the cows are mixed together generated by DF-GAN.
The images in the 6th - 8th columns are poorly synthesized by DM-GAN: the objects cannot be recognized and the backgrounds are fuzzy.
In the 7th and 8th columns, the ``skier'' and ``elephants" generated by DF-GAN do not seem as a natural part in the corresponding image.
These qualitative examples on the more challenging COCO dataset demonstrate that SSA-GAN is able to generate a complex image with multiple objects as well as the corresponding background.

\subsection{Ablation Studies}
\label{subsec:ablation}
In this subsection, we verify the effectiveness of each component in SSA-GAN by conducting  extensive ablation studies on the testing set of the CUB dataset~\cite{wah2011caltech}.

\begin{figure*}[t!]
\centering
\hspace{-1cm}
\begin{minipage}[c]{0.24\textwidth}
\center{GT}
\end{minipage}
\hfill
\hspace{-1.5cm}
\begin{minipage}[c]{0.24\textwidth}
\center{Input: A close up of a boat on a field with a cloudy sky.}
\end{minipage}
\hfill
\begin{minipage}[c]{0.24\textwidth}
\center{A close up of a boat on a field \textbf{under a sunset}.}
\end{minipage}
\hfill
\begin{minipage}[c]{0.24\textwidth}
\center{A close up of a boat on a field with \textbf{a clear sky}.}
\end{minipage}

\begin{minipage}{0.13\textwidth}
\includegraphics[width=\textwidth]{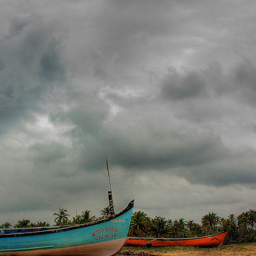}
\end{minipage}
\hfill
\begin{minipage}{0.13\textwidth}
\includegraphics[width=\textwidth]{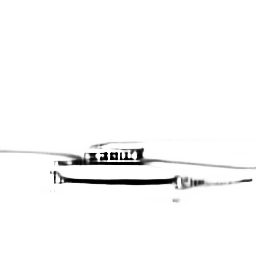}
\end{minipage}
\hfill
\begin{minipage}{0.13\textwidth}
\includegraphics[width=\textwidth]{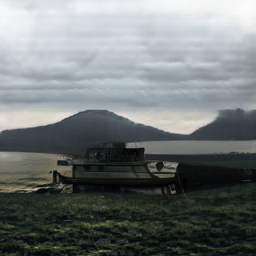}
\end{minipage}
\hfill
\begin{minipage}{0.13\textwidth}
\includegraphics[width=\textwidth]{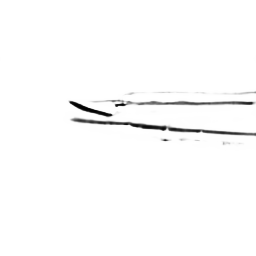}
\end{minipage}
\hfill
\begin{minipage}{0.13\textwidth}
\includegraphics[width=\textwidth]{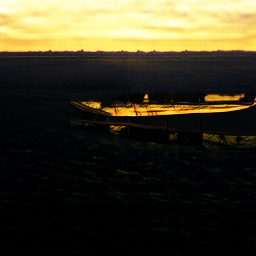}
\end{minipage}
\hfill
\begin{minipage}{0.13\textwidth}
\includegraphics[width=\textwidth]{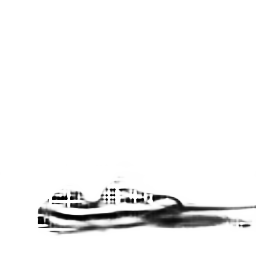}
\end{minipage}
\hfill
\begin{minipage}{0.13\textwidth}
\includegraphics[width=\textwidth]{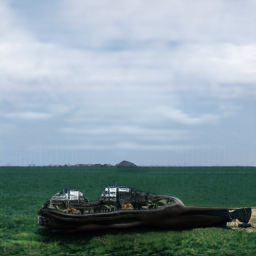}
\end{minipage}

\hspace{-1cm}
\begin{minipage}[c]{0.24\textwidth}
 \center{GT}
\end{minipage}
\hfill
\hspace{-1.5cm}
\begin{minipage}[c]{0.24\textwidth}
\center{Input: A person skiing down a snow covered mountain.}
\end{minipage}
\hfill
\begin{minipage}[c]{0.24\textwidth}
\center{\textbf{Two men} skiing down a snow covered mountain.}
\end{minipage}
\hfill
\begin{minipage}[c]{0.24\textwidth}
\center{A person \textbf{walking} down a \textbf{grass} covered mountain.}
\end{minipage}

\begin{minipage}{0.13\textwidth}
\includegraphics[width=\textwidth]{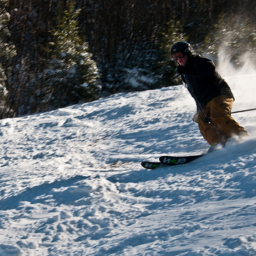}
\end{minipage}
\hfill
\begin{minipage}{0.13\textwidth}
\includegraphics[width=\textwidth]{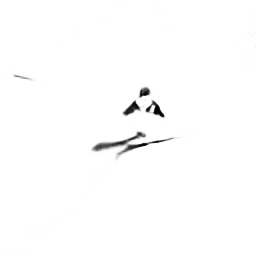}
\end{minipage}
\hfill
\begin{minipage}{0.13\textwidth}
\includegraphics[width=\textwidth]{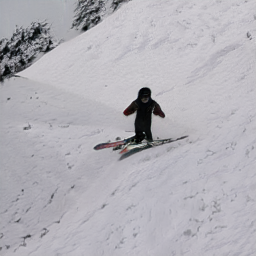}
\end{minipage}
\hfill
\begin{minipage}{0.13\textwidth}
\includegraphics[width=\textwidth]{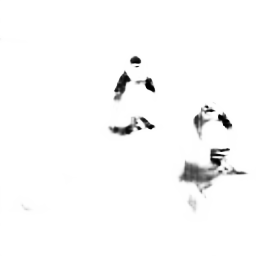}
\end{minipage}
\hfill
\begin{minipage}{0.134\textwidth}
\includegraphics[width=\textwidth]{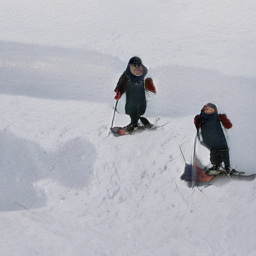}
\end{minipage}
\hfill
\begin{minipage}{0.13\textwidth}
\includegraphics[width=\textwidth]{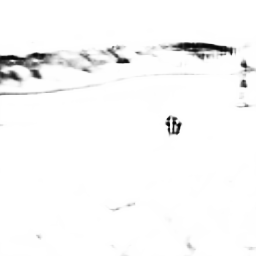}
\end{minipage}
\hfill
\begin{minipage}{0.13\textwidth}
\includegraphics[width=\textwidth]{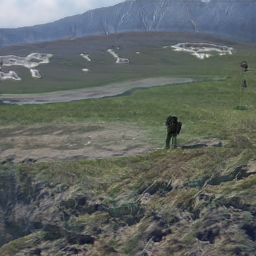}
\end{minipage}
\caption{Examples of diverse image generation by changing some words in the input text (in \textbf{bold}) on the test set of the COCO dataset. The predicted mask for each generated image is also shown (on its left side). Best view in color and zoom in.}
\label{fig:coco_change}
\vspace{-4mm}
\end{figure*}

\vspace{-4mm}
\paragraph{SSACN and DAMSM} Firstly, we verify how the proposed SSACN block and the additional DAMSM affect the performance of the network. The results of using different components are given in Table~\ref{tab:ablation_study_components}. We treat the DF-GAN as the baseline denoted (ID0). Replacing the UPBlocks in DF-GAN with our SSACN blocks, both the IS and FID performance are improved (ID1), which shows that our SSACN block is able to fuse text and image features better. When DAMSM is added to our network (ID2), the overall performance is improved. 
It indicates that DAMSM helps improve the text-image consistency. Then, we train the whole framework in order to fine tune the text encoder (ID3). Our method achieves further improvements in IS but inferior performance in FID. The reason is that fine tuning the text encoder helps text-image fusion and improves the text-image consistency so that the IS score is improved. However, when the encoded text features become more adaptive to the image features, the diversity of generated images also increases (more deeply constrained by the diverse text descriptions). Thus, the FID performance decreases while it measures the KL divergence between the real images and generated images.
It is worth noting that, without adding DAMSM, our method (ID1) achieves better performance   compared to the most recent state-of-the-art method DF-GAN \cite{ming2020DFGAN} (ID0).

\begin{table}[t!]
\caption{Ablation study of evaluating how the performance is affected by different numbers of mask maps used in the SSA-GAN. Note that, text encoder is not fine tuned here.}
\label{tab:results_num_masks}
\vspace{-4mm}
\begin{center}
\begin{tabular}{c c c c}
\hline
Parameter & Stages & IS $\uparrow$ & FID $\downarrow$ \\
\hline
\multirow{6}{*}{\#masks} & 2 & 4.98 $\pm$ 0.09 & 19.69 \\ 
& 3 & 5.04 $\pm$ 0.07 & 18.40 \\
& 4 & 5.05 $\pm$ 0.05 & \textbf{15.03} \\
& 5 & 5.02 $\pm$ 0.07 & 17.64 \\
& 6 & 4.97 $\pm$ 0.04 & 16.62 \\
& 7 & \textbf{5.07 $\pm$ 0.04} & 15.61 \\
\hline
\end{tabular}  
\end{center}
\vspace{-5mm}
\end{table}

\vspace{-3mm}
\paragraph{Mask Maps}
The predicted mask maps provide spatial information for the our SSCBN. To evaluate how the mask maps affect the text-image fusion process, we add the mask predictor one by one from the last SSACN block to the first one and observe how the performance varies. The results are given in Table~\ref{tab:results_num_masks}. 
We can see that the performance increases constantly by increasing the mask maps up to 4. However, the performance is marginally worse when adding the 5th and 6th mask maps. When the framework uses 7 mask maps, it has the highest IS score and second best FID performance. 
This phenomenon demonstrates that more mask maps help text-image fusion process, so that the generated images are more realistic and text-image consistent (higher IS scores). Meanwhile, deeper text-image fusion also makes the generated images be stronger controlled by the diverse text descriptions.
Consequently, the generated images become more diverse, which leads to higher FID.
Note that, we use 7 mask maps for all the rest experiments in this work.

To gain more insight, Fig.~\ref{fig:cub_mask_layers} shows the mask maps learned on different stages. We can see that, the mask maps become more focused on the bird when the text-image fusion becomes deeper. Especially in the last two stages, the main attention is on the whole bird to generate the bird, then on the specific local parts of the bird in order to refine the details of the bird. It visually demonstrates that the mask maps are predicted based on the current generated image features and deepen the text-image fusion process.

\vspace{-2mm}
\paragraph{Diverse Image Generation}
We conduct the experiments to show the ability of generating image by changing some words in the text description.
We further conduct an ablation study by modifying some words in the given text, in order to evaluate the ability of generating diverse images and keeping the semantic text-image consistency.
The qualitative results are shown in Fig.~\ref{fig:cub_change_color} and Fig.~\ref{fig:coco_change}.
From Fig.~\ref{fig:cub_change_color}, we can see that, the color of the generated bird changes in order to keep consistent with the specific text conditions.
In Fig.~\ref{fig:coco_change}, the generated images are also semantic consistent with the modified text.
It is worth noting that, the predicted mask is a surprising high-quality sketch corresponding to the text description, especially for the bird generation.

\section{Conclusion}

In this paper, we proposed a novel framework of Semantic-Spatial Aware GAN (SSA-GAN) for T2I generation. 
It has one generator-discriminator pair and is trained end-to-end. 
The core module of SSA-GAN is a  Semantic-Spatial  Aware  Convolution  Network  (SSACN) block which operates Semantic-Spatial Condition Batch Normalization by predicting mask maps based on the current generated image features, and learning the affine parameters from the encoded text vector. 
The SSACN block deepens the text-image fusion through the image generation process, and guarantees the text-image consistency.
In our  experimental results and ablation studies, we demonstrated the effectiveness of our proposed model and significant improvement over previous state-of-the-art in terms of T2I generation.

\section*{Appendix}
\noindent
In this Appendix, we provide more qualitative examples 
for further discussion in Sec.~\ref{sec:qualitative}. 
For the purpose of completeness,  we briefly give a review on DAMSM Loss~\cite{xu2018attngan} in Sec.~\ref{sec:damsm}.
Code is available at \url{https://github.com/wtliao/text2image}.


\begin{figure*}[t!]
\centering

\begin{minipage}[c]{0.01\textwidth}
\fontsize{2.0pt}{0.5\baselineskip}\selectfont \center{\ } 
\end{minipage}
\hfill
\begin{minipage}[t]{0.115\textwidth}
\center{\footnotesize{This bird has a white and gray speckled belly and breast with a black eyering and short bill.}}
\end{minipage}
\hfill
\begin{minipage}[t]{0.115\textwidth}
\center{\footnotesize{This bird has a grey head, a short flat beak and long legs.}}
\end{minipage}
\hfill
\begin{minipage}[t]{0.115\textwidth}
\center{\footnotesize{This long billed water bird has white body and an orange bill.}} 
\end{minipage}
\hfill
\begin{minipage}[t]{0.115\textwidth}
\center{\footnotesize{This bird is brown and yellow in color with a brown beak.}}
\end{minipage}
\hfill
\begin{minipage}[t]{0.115\textwidth}
\center{\footnotesize{This bird is white and red in color with a brown beak and dark eye rings.}}
\end{minipage}
\hfill
\begin{minipage}[t]{0.115\textwidth}
\center{\footnotesize{This bird has a black stripe over its eyes and a grey belly.}}
\end{minipage}
\hfill
\begin{minipage}[t]{0.115\textwidth}
\center{\footnotesize{This little bird has a white belly and breast with black and white wings a yellow on its crown.}}
\end{minipage}
\hfill
\begin{minipage}[t]{0.115\textwidth}
\center{\footnotesize{The bird is small gray has a red crown and a small yellow bill.}}
\end{minipage}
\vspace{2pt}

\begin{minipage}[c]{0.01\textwidth}
\center{\rotatebox{90}{GT}}
\end{minipage}
\hfill
\begin{minipage}{0.115\textwidth}
\includegraphics[width=\textwidth]{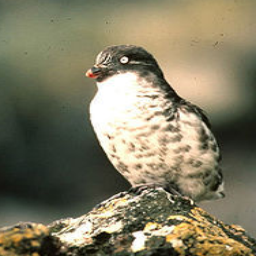}
\end{minipage}
\hfill
\begin{minipage}{0.115\textwidth}
\includegraphics[width=\textwidth]{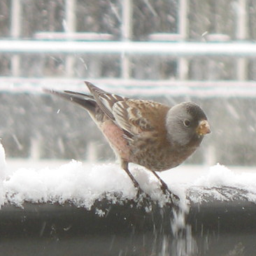}
\end{minipage}
\hfill
\begin{minipage}{0.115\textwidth}
\includegraphics[width=\textwidth]{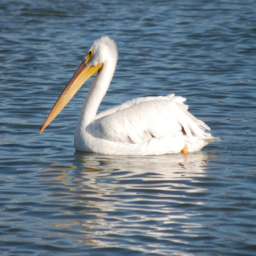}
\end{minipage}
\hfill
\begin{minipage}{0.115\textwidth}
\includegraphics[width=\textwidth]{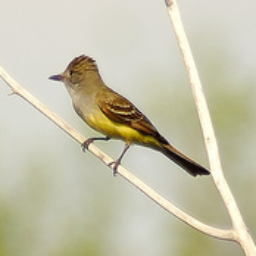}
\end{minipage}
\hfill
\begin{minipage}{0.115\textwidth}
\includegraphics[width=\textwidth]{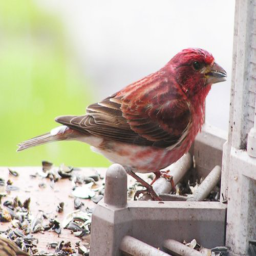}
\end{minipage}
\hfill
\begin{minipage}{0.115\textwidth}
\includegraphics[width=\textwidth]{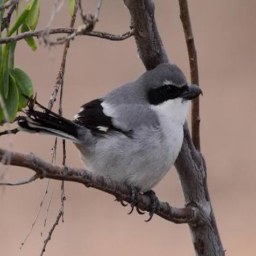}
\end{minipage}
\hfill
\begin{minipage}{0.115\textwidth}
\includegraphics[width=\textwidth]{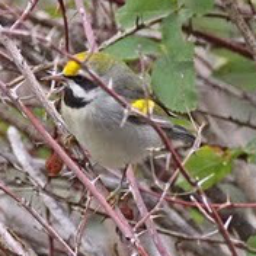}
\end{minipage}
\hfill
\begin{minipage}{0.115\textwidth}
\includegraphics[width=\textwidth]{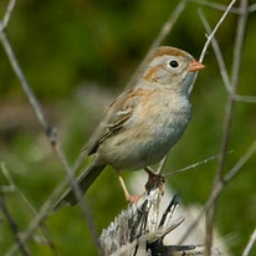}
\end{minipage}
\vspace{2pt}

\begin{minipage}[c]{0.01\textwidth}
\center{\rotatebox{90}{DM-GAN}}
\end{minipage}
\hfill
\begin{minipage}{0.115\textwidth}
\includegraphics[width=\textwidth]{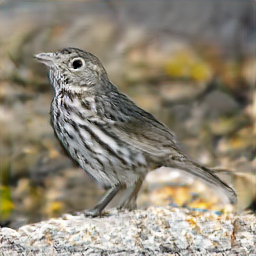}
\end{minipage}
\hfill
\begin{minipage}{0.115\textwidth}
\includegraphics[width=\textwidth]{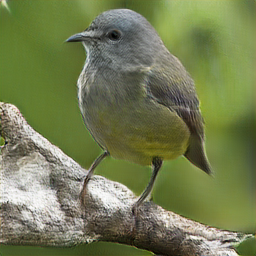}
\end{minipage}
\hfill
\begin{minipage}{0.115\textwidth}
\includegraphics[width=\textwidth]{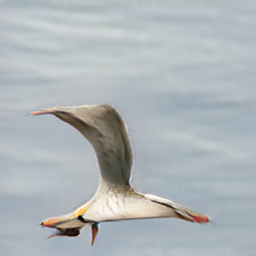}
\end{minipage}
\hfill
\begin{minipage}{0.115\textwidth}
\includegraphics[width=\textwidth]{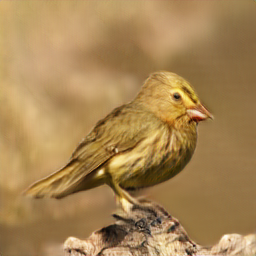}
\end{minipage}
\hfill
\begin{minipage}{0.115\textwidth}
\includegraphics[width=\textwidth]{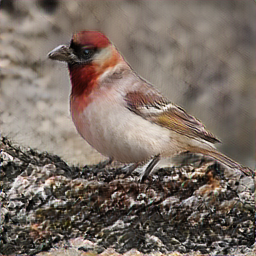}
\end{minipage}
\hfill
\begin{minipage}{0.115\textwidth}
\includegraphics[width=\textwidth]{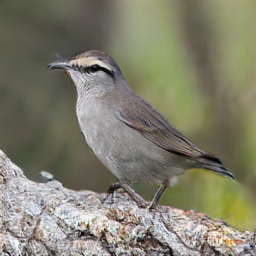}
\end{minipage}
\hfill
\begin{minipage}{0.115\textwidth}
\includegraphics[width=\textwidth]{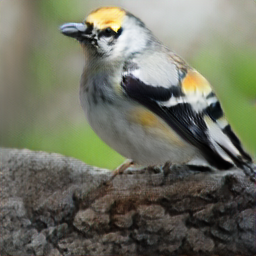}
\end{minipage}
\hfill
\begin{minipage}{0.115\textwidth}
\includegraphics[width=\textwidth]{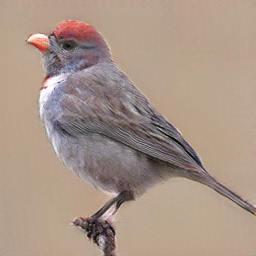}
\end{minipage}
\vspace{2pt}

\begin{minipage}[c]{0.01\textwidth}
\center{\rotatebox{90}{DF-GAN}}
\end{minipage}
\hfill
\begin{minipage}{0.115\textwidth}
\includegraphics[width=\textwidth]{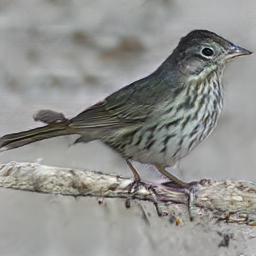}
\end{minipage}
\hfill
\begin{minipage}{0.115\textwidth}
\includegraphics[width=\textwidth]{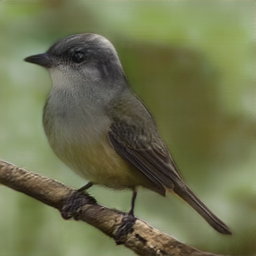}
\end{minipage}
\hfill
\begin{minipage}{0.115\textwidth}
\includegraphics[width=\textwidth]{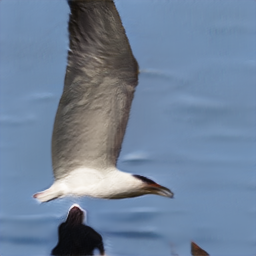}
\end{minipage}
\hfill
\begin{minipage}{0.115\textwidth}
\includegraphics[width=\textwidth]{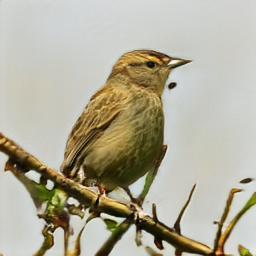}
\end{minipage}
\hfill
\begin{minipage}{0.115\textwidth}
\includegraphics[width=\textwidth]{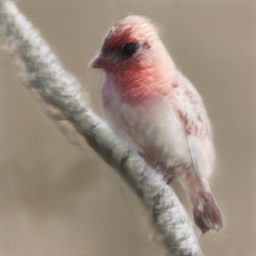}
\end{minipage}
\hfill
\begin{minipage}{0.115\textwidth}
\includegraphics[width=\textwidth]{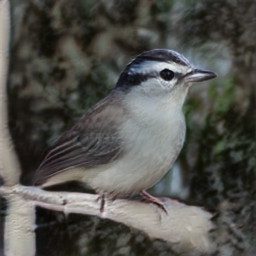}
\end{minipage}
\hfill
\begin{minipage}{0.115\textwidth}
\includegraphics[width=\textwidth]{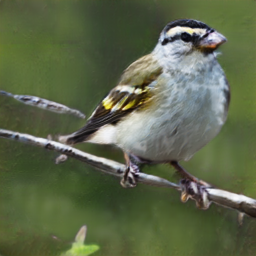}
\end{minipage}
\hfill
\begin{minipage}{0.115\textwidth}
\includegraphics[width=\textwidth]{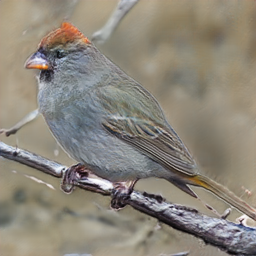}
\end{minipage}
\vspace{2pt}

\begin{minipage}[c]{0.01\textwidth}
\center{\rotatebox{90}{Ours}}
\end{minipage}
\hfill
\begin{minipage}{0.115\textwidth}
\includegraphics[width=\textwidth]{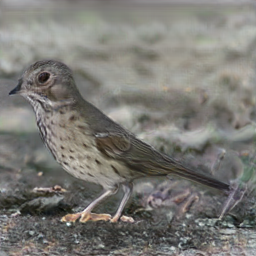}
\end{minipage}
\hfill
\begin{minipage}{0.115\textwidth}
\includegraphics[width=\textwidth]{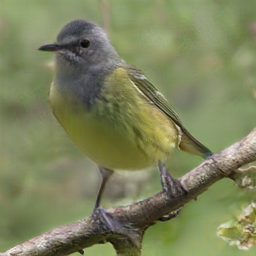}
\end{minipage}
\hfill
\begin{minipage}{0.115\textwidth}
\includegraphics[width=\textwidth]{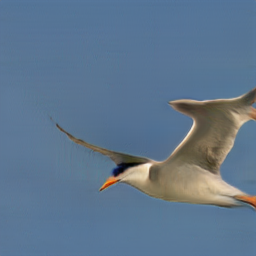}
\end{minipage}
\hfill
\begin{minipage}{0.115\textwidth}
\includegraphics[width=\textwidth]{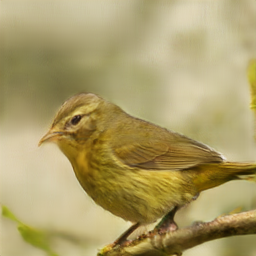}
\end{minipage}
\hfill
\begin{minipage}{0.115\textwidth}
\includegraphics[width=\textwidth]{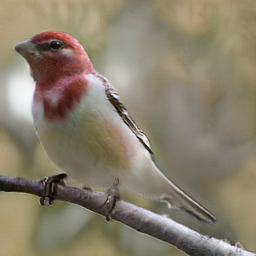}
\end{minipage}
\hfill
\begin{minipage}{0.115\textwidth}
\includegraphics[width=\textwidth]{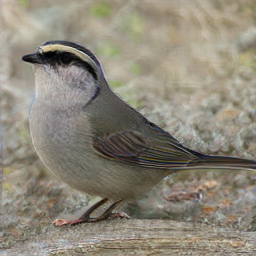}
\end{minipage}
\hfill
\begin{minipage}{0.115\textwidth}
\includegraphics[width=\textwidth]{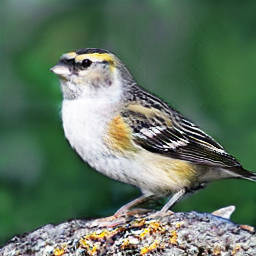}
\end{minipage}
\hfill
\begin{minipage}{0.115\textwidth}
\includegraphics[width=\textwidth]{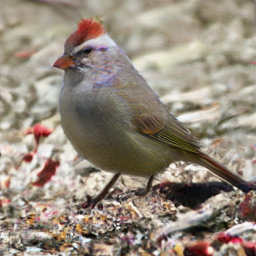}
\end{minipage}
\vspace{2pt}

\caption{Qualitative comparison between our method and the recent state-of-the-art methods DM-GAN \cite{zhu2019dm}, DF-GAN \cite{ming2020DFGAN} on the test set of CUB bird dataset. The input text descriptions are given in the first row and the corresponding generated images from different methods are shown in the same column. Best view in color and zoom in.}
\label{fig:qualitative_cub}
\end{figure*}

\section*{Qualitative Examples}
\label{sec:qualitative}
\begin{figure*}[t!]
\centering

\begin{minipage}[c]{0.01\textwidth}
\selectfont {\ } 
\end{minipage}
\hfill
\begin{minipage}[t]{0.11\textwidth}
\center{\footnotesize{A man riding skis down a snow covered slope.}} 
\end{minipage}
\hfill
\begin{minipage}[t]{0.155\textwidth}
\center{\footnotesize{A slightly overcooked homemade personal sized pizza with meat and red peppers.}} 
\end{minipage}
\hfill
\begin{minipage}[t]{0.155\textwidth}
\center{\footnotesize{A group of people are in a green field.}} 
\end{minipage}
\hfill
\begin{minipage}[t]{0.155\textwidth}
\center{\footnotesize{The surfer is riding a wave into shore.}} 
\end{minipage}
\hfill
\begin{minipage}[t]{0.11\textwidth}
\center{\footnotesize{Some horses in a field of green grass with a sky in the background.}}
\end{minipage}
\hfill
\begin{minipage}[t]{0.155\textwidth}
\center{\footnotesize{The cabin is very clean and empty filled with wood.}}
\end{minipage}
\vspace{2pt}

\begin{minipage}[c]{0.01\textwidth}
\center{\rotatebox{90}{GT}}
\end{minipage}
\hfill
\begin{minipage}{0.155\textwidth}
\includegraphics[width=\textwidth]{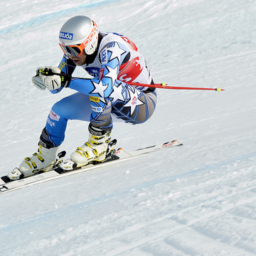}
\end{minipage}
\hfill
\begin{minipage}{0.155\textwidth}
\includegraphics[width=\textwidth]{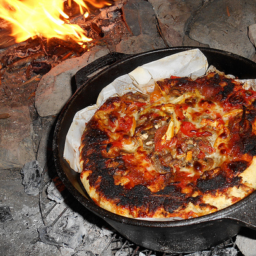}
\end{minipage}
\hfill
\begin{minipage}{0.155\textwidth}
\includegraphics[width=\textwidth]{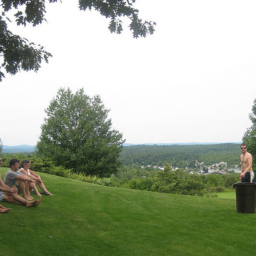}
\end{minipage}
\hfill
\begin{minipage}{0.155\textwidth}
\includegraphics[width=\textwidth]{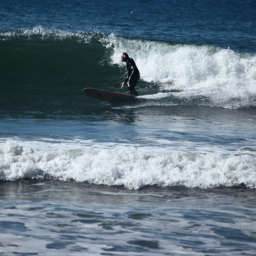}
\end{minipage}
\hfill
\begin{minipage}{0.155\textwidth}
\includegraphics[width=\textwidth]{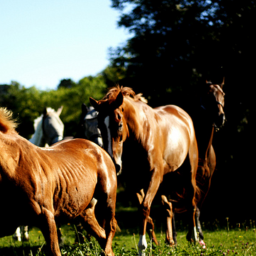}
\end{minipage}
\hfill
\begin{minipage}{0.155\textwidth}
\includegraphics[width=\textwidth]{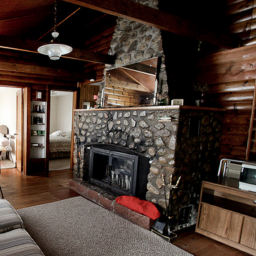}
\end{minipage}
\vspace{2pt}

\begin{minipage}[c]{0.01\textwidth}
\center{\rotatebox{90}{DM-GAN}}
\end{minipage}
\hfill
\begin{minipage}{0.155\textwidth}
\includegraphics[width=\textwidth]{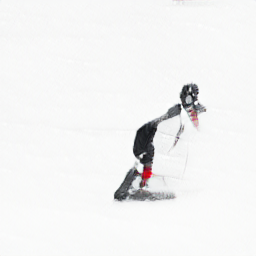}
\end{minipage}
\hfill
\begin{minipage}{0.155\textwidth}
\includegraphics[width=\textwidth]{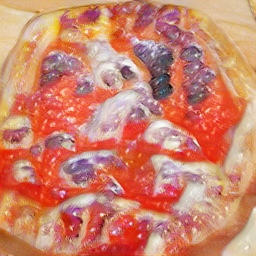}
\end{minipage}
\hfill
\begin{minipage}{0.155\textwidth}
\includegraphics[width=\textwidth]{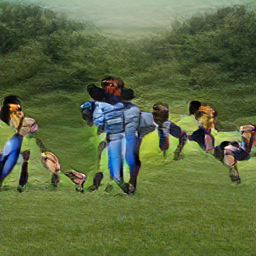}
\end{minipage}
\hfill
\begin{minipage}{0.155\textwidth}
\includegraphics[width=\textwidth]{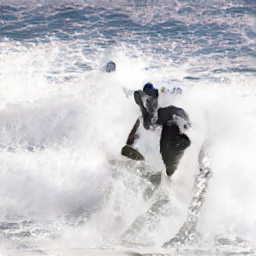}
\end{minipage}
\hfill
\begin{minipage}{0.155\textwidth}
\includegraphics[width=\textwidth]{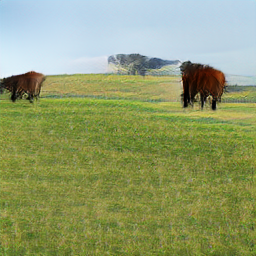}
\end{minipage}
\hfill
\begin{minipage}{0.155\textwidth}
\includegraphics[width=\textwidth]{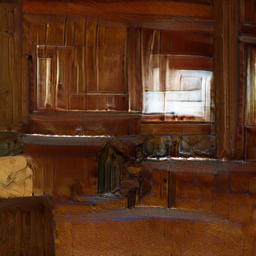}
\end{minipage}
\vspace{2pt}

\begin{minipage}[c]{0.01\textwidth}
\center{\rotatebox{90}{DF-GAN}}
\end{minipage}
\hfill
\begin{minipage}{0.155\textwidth}
\includegraphics[width=\textwidth]{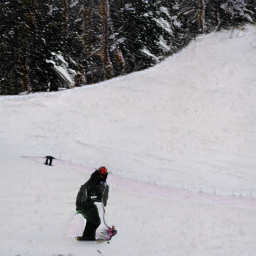}
\end{minipage}
\hfill
\begin{minipage}{0.155\textwidth}
\includegraphics[width=\textwidth]{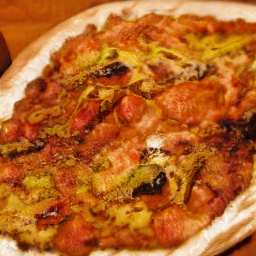}
\end{minipage}
\hfill
\begin{minipage}{0.155\textwidth}
\includegraphics[width=\textwidth]{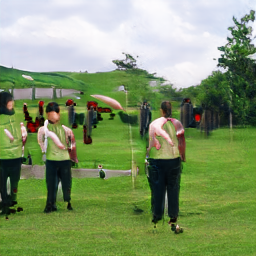}
\end{minipage}
\hfill
\begin{minipage}{0.155\textwidth}
\includegraphics[width=\textwidth]{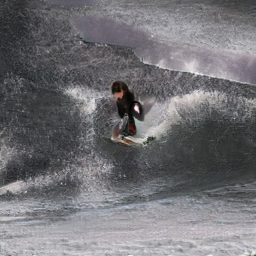}
\end{minipage}
\hfill
\begin{minipage}{0.155\textwidth}
\includegraphics[width=\textwidth]{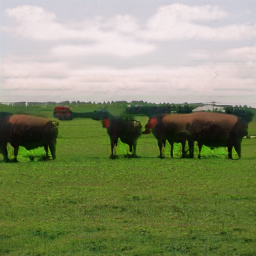}
\end{minipage}
\hfill
\begin{minipage}{0.155\textwidth}
\includegraphics[width=\textwidth]{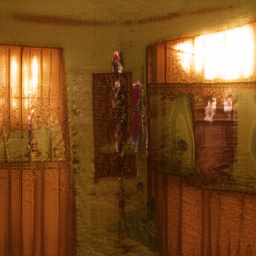}
\end{minipage}
\vspace{2pt}

\begin{minipage}[c]{0.01\textwidth}
\center{\rotatebox{90}{Ours}}
\end{minipage}
\hfill
\begin{minipage}{0.155\textwidth}
\includegraphics[width=\textwidth]{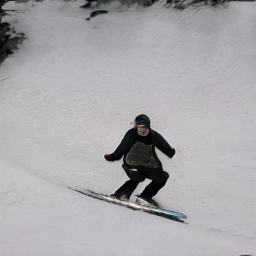}
\end{minipage}
\hfill
\begin{minipage}{0.155\textwidth}
\includegraphics[width=\textwidth]{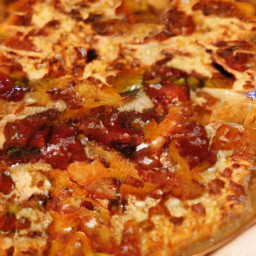}
\end{minipage}
\hfill
\begin{minipage}{0.155\textwidth}
\includegraphics[width=\textwidth]{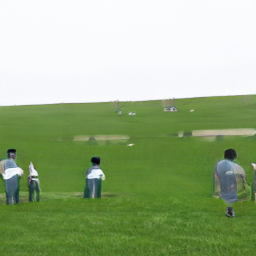}
\end{minipage}
\hfill
\begin{minipage}{0.155\textwidth}
\includegraphics[width=\textwidth]{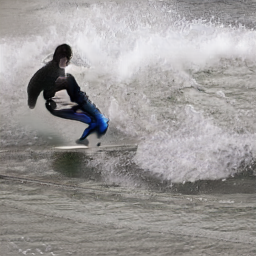}
\end{minipage}
\hfill
\begin{minipage}{0.155\textwidth}
\includegraphics[width=\textwidth]{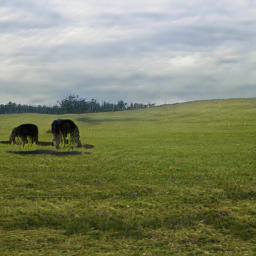}
\end{minipage}
\hfill
\begin{minipage}{0.155\textwidth}
\includegraphics[width=\textwidth]{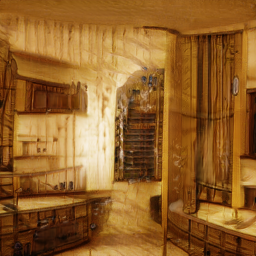}
\end{minipage}
\vspace{2pt}

\caption{Qualitative comparison between our method and the recent state-of-the-art methods DM-GAN \cite{zhu2019dm}, DF-GAN \cite{ming2020DFGAN} on the test set of COCO dataset. The input text descriptions are given in the first row and the corresponding generated images from different methods are shown in the same column. Best view in color and zoom in.}
\label{fig:qualitative_coco}
\end{figure*}

More qualitative examples are provided on CUB (Fig.~\ref{fig:qualitative_cub}) and COCO (Fig.~\ref{fig:qualitative_coco}) datasets respectively. The CUB bird dataset focuses on synthesizing the details of a bird while the COCO dataset focuses on synthesizing multiple objects with various backgrounds.

Fig.~\ref{fig:qualitative_cub}, one can observe that the birds generated by our methods are more vivid and better match the attributes described in the given text compared to the other methods. For example in the 3rd column, the ``orange bill" is not generated by other methods and the whole birds synthesized by other methods seem not real, while our method generates a bird that has all attributes mentioned in the text and likes a real one.

Fig.~\ref{fig:qualitative_coco} demonstrates that, our method is able to generate more realistic and better complex images that have multiple objects and various backgrounds from text. Take the first column as an example, neither the ``man'' nor the background is generated by DM-GAN well. The skier generated by DF-GAN seems not real. 

\begin{figure*}[t!]
\centering
\begin{minipage}[c]{0.15\textwidth}
\center{Input Text} 
\end{minipage}
\hfill
\begin{minipage}[t]{0.10\textwidth}
\center{Ours}
\end{minipage}
\hfill
\begin{minipage}[t]{0.10\textwidth}
\center{1}
\end{minipage}
\hfill
\begin{minipage}[t]{0.10\textwidth}
\center{2} 
\end{minipage}
\hfill
\begin{minipage}[t]{0.10\textwidth}
\center{3}
\end{minipage}
\hfill
\begin{minipage}[t]{0.10\textwidth}
\center{4}
\end{minipage}
\hfill
\begin{minipage}[t]{0.10\textwidth}
\center{5}
\end{minipage}
\hfill
\begin{minipage}[t]{0.10\textwidth}
\center{6}
\end{minipage}
\hfill
\begin{minipage}[t]{0.10\textwidth}
\center{7}
\end{minipage}
\vspace{2pt}

\begin{minipage}{0.15\textwidth}
\begin{small}
A small bird with a brown and red coloring.
\end{small}
\end{minipage}
\begin{minipage}{0.84\textwidth}
\includegraphics[width=0.12\textwidth]{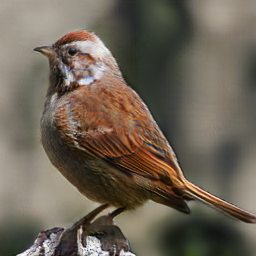}
\includegraphics[width=0.12\textwidth]{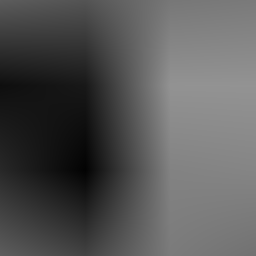}
\includegraphics[width=0.12\textwidth]{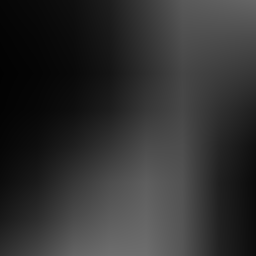}
\includegraphics[width=0.12\textwidth]{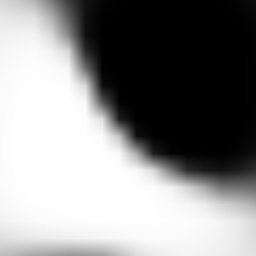}
\includegraphics[width=0.12\textwidth]{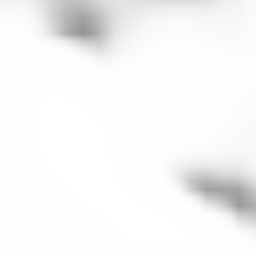}
\includegraphics[width=0.12\textwidth]{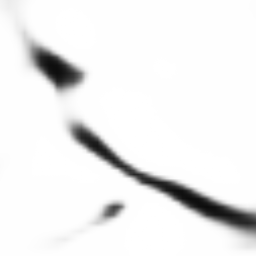}
\includegraphics[width=0.12\textwidth]{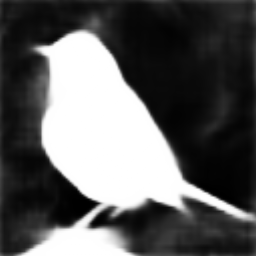}
\includegraphics[width=0.12\textwidth]{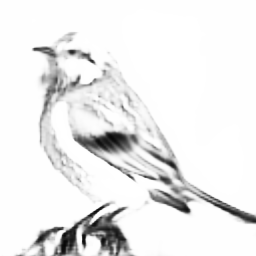}
\end{minipage}
\vfill

\begin{minipage}{0.15\textwidth}
\begin{small}
This is a small grey bird with brown wings and a small black beak.
\end{small}
\end{minipage}
\begin{minipage}{0.84\textwidth}
\includegraphics[width=0.12\textwidth]{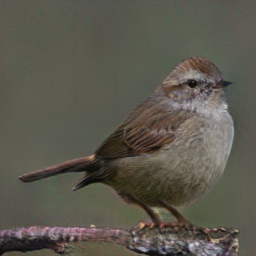}
\includegraphics[width=0.12\textwidth]{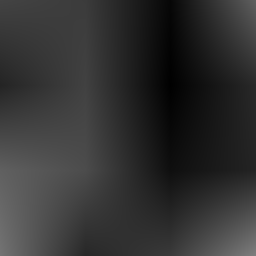}
\includegraphics[width=0.12\textwidth]{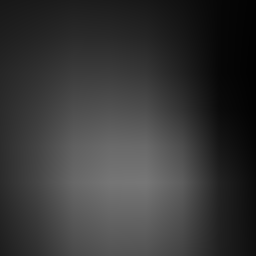}
\includegraphics[width=0.12\textwidth]{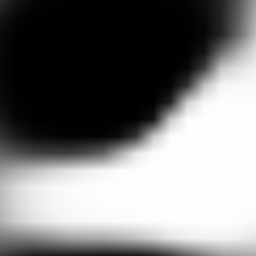}
\includegraphics[width=0.12\textwidth]{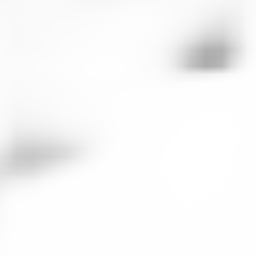}
\includegraphics[width=0.12\textwidth]{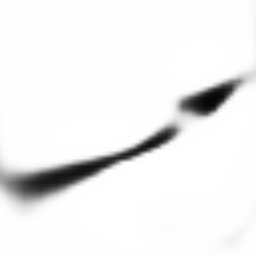}
\includegraphics[width=0.12\textwidth]{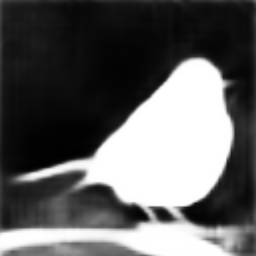}
\includegraphics[width=0.12\textwidth]{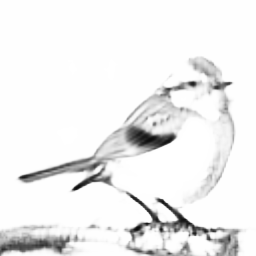}
\end{minipage}
\vfill 

\begin{minipage}{0.15\textwidth}
\begin{small}
A herd of black and white cattle standing on a field.
\end{small}
\end{minipage}
\begin{minipage}{0.84\textwidth}
\includegraphics[width=0.12\textwidth]{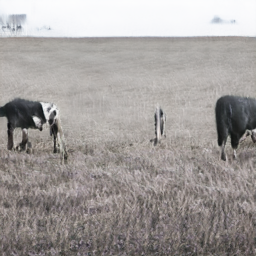}
\includegraphics[width=0.12\textwidth]{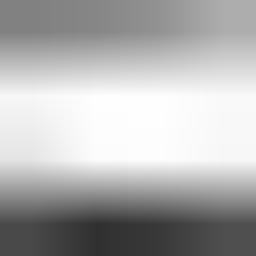}
\includegraphics[width=0.12\textwidth]{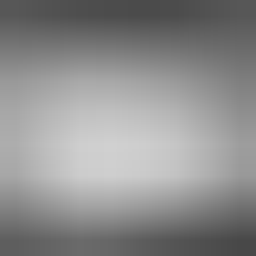}
\includegraphics[width=0.12\textwidth]{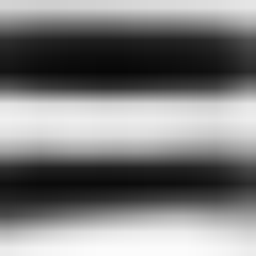}
\includegraphics[width=0.12\textwidth]{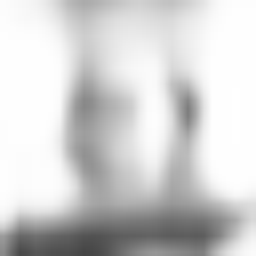}
\includegraphics[width=0.12\textwidth]{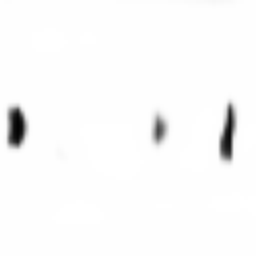}
\includegraphics[width=0.12\textwidth]{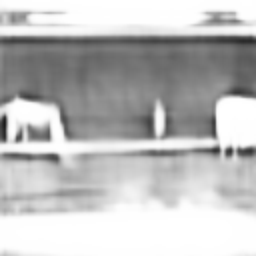}
\includegraphics[width=0.12\textwidth]{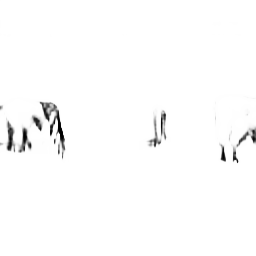}
\end{minipage}
\vfill

\begin{minipage}{0.15\textwidth}
\begin{small}
A table top with some trey of food on it.
\end{small}
\end{minipage}
\begin{minipage}{0.84\textwidth}
\includegraphics[width=0.12\textwidth]{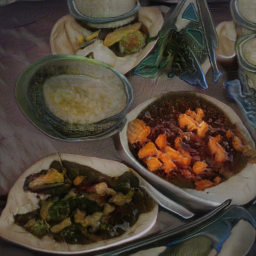}
\includegraphics[width=0.12\textwidth]{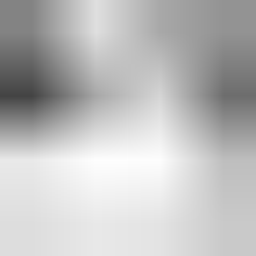}
\includegraphics[width=0.12\textwidth]{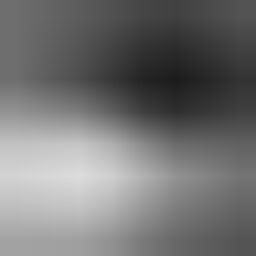}
\includegraphics[width=0.12\textwidth]{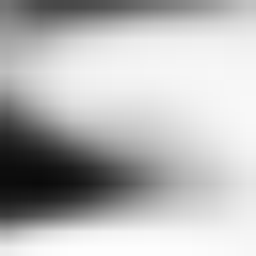}
\includegraphics[width=0.12\textwidth]{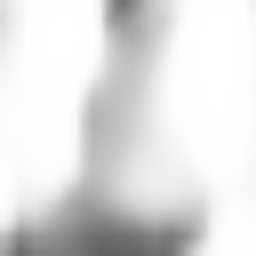}
\includegraphics[width=0.12\textwidth]{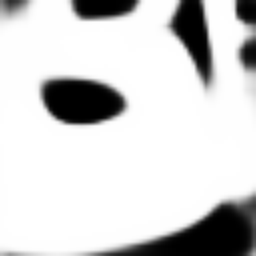}
\includegraphics[width=0.12\textwidth]{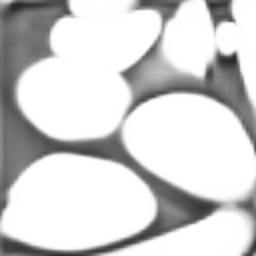}
\includegraphics[width=0.12\textwidth]{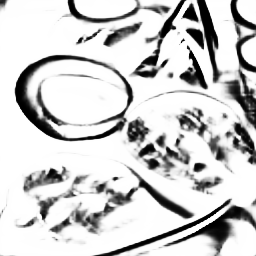}
\end{minipage}
\caption{Examples of mask maps predicted in different SSACN blocks (indicated by the numbers in the first row). Best view in color and zoom in.}
\label{fig:mask_layers}
\vspace{1mm}
\end{figure*}
\subsection*{Mask Prediction}
Fig.~\ref{fig:mask_layers} shows some examples of predicted mask maps from different SSACN stages on CUB (first two rows) and COCO (last two rows) datasets, respectively. From left to right of the mask maps, one can observe that, in the 1st and 2nd stages, the predicted mask maps do not obviously indicate where to fuse the text information on the image feature maps. Because the text-image fusion is too shallow and the whole image feature maps require more text information. From the 3rd stage, more attention is paid to the attributes or objects that the text descriptions have mentioned. 
Due to the deeper text-image fusion process, our proposed SSACN block is able to predict which part of the current image feature maps needs to be refined with the text information. In particular, the predicted mask maps pay the main attention to the rough layout and background of the generated image (5th stage), then focus on generating individual objects (6th stage) and finally concentrate on the details of each objects (7th stage). The process of predicting the mask maps reveals that  the SSACN block is able to (1) precisely predict where of the image feature maps need to be refined by the text information based on the current generated image features, and (2) effectively deepen the text-image fusion process through the image generation process.

\begin{figure*}[t!]
\centering
\begin{minipage}{\textwidth}
\center{\textbf{Input text:} A colorful $<$\textbf{color}$>$ bird has wings with dark stripes and small eyes.}
\end{minipage}
\vfill

\begin{minipage}{0.05\textwidth}
\center{\small{blue}}
\end{minipage}
\hfill
\begin{minipage}{0.11\textwidth}
\includegraphics[width=\textwidth]{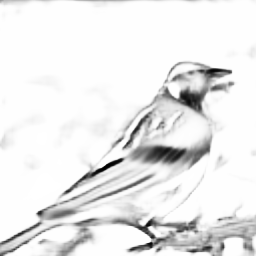}
\end{minipage}
\hfill
\begin{minipage}{0.11\textwidth}
\includegraphics[width=\textwidth]{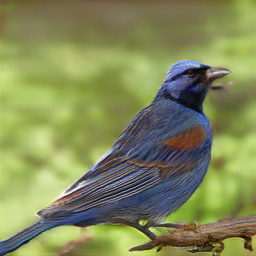}
\end{minipage}
\hfill
\begin{minipage}{0.11\textwidth}
\includegraphics[width=\textwidth]{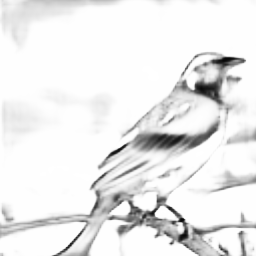}
\end{minipage}
\hfill
\begin{minipage}{0.11\textwidth}
\includegraphics[width=\textwidth]{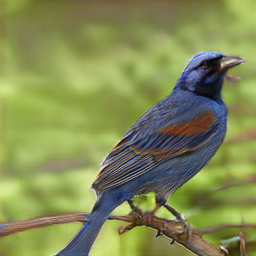}
\end{minipage}
\hfill
\begin{minipage}{0.11\textwidth}
\includegraphics[width=\textwidth]{./fig/fig3/mask_117.png}
\end{minipage}
\hfill
\begin{minipage}{0.11\textwidth}
\includegraphics[width=\textwidth]{./fig/fig3/image_117.png}
\end{minipage}
\hfill
\begin{minipage}{0.11\textwidth}
\includegraphics[width=\textwidth]{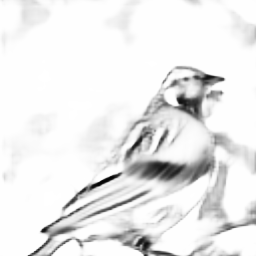}
\end{minipage}
\hfill
\begin{minipage}{0.11\textwidth}
\includegraphics[width=\textwidth]{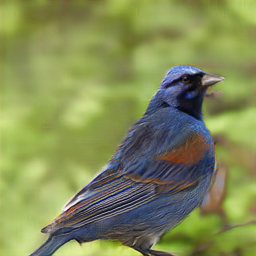}
\end{minipage}
\vfill

\begin{minipage}{0.05\textwidth}
\center{\small{red}}
\end{minipage}
\hfill
\begin{minipage}{0.11\textwidth}
\includegraphics[width=\textwidth]{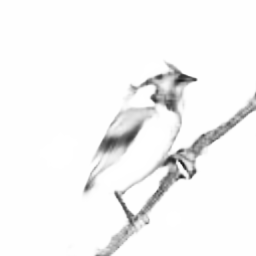}
\end{minipage}
\hfill
\begin{minipage}{0.11\textwidth}
\includegraphics[width=\textwidth]{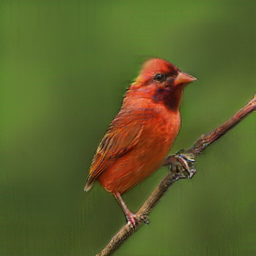}
\end{minipage}
\hfill
\begin{minipage}{0.11\textwidth}
\includegraphics[width=\textwidth]{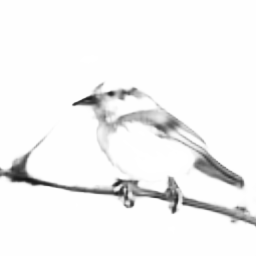}
\end{minipage}
\hfill
\begin{minipage}{0.11\textwidth}
\includegraphics[width=\textwidth]{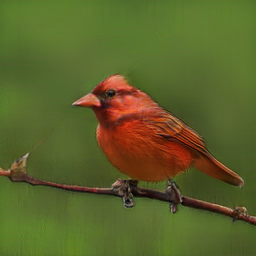}
\end{minipage}
\hfill
\begin{minipage}{0.11\textwidth}
\includegraphics[width=\textwidth]{./fig/fig3/mask_38.png}
\end{minipage}
\hfill
\begin{minipage}{0.11\textwidth}
\includegraphics[width=\textwidth]{./fig/fig3/image_38.png}
\end{minipage}
\hfill
\begin{minipage}{0.11\textwidth}
\includegraphics[width=\textwidth]{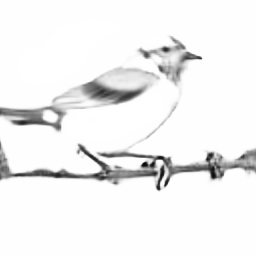}
\end{minipage}
\hfill
\begin{minipage}{0.11\textwidth}
\includegraphics[width=\textwidth]{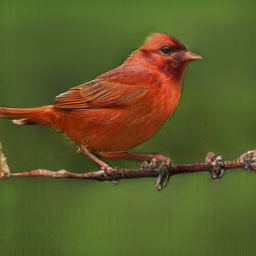}
\end{minipage}
\vfill

\begin{minipage}{0.05\textwidth}
\center{\small{white}}
\end{minipage}
\hfill
\begin{minipage}{0.11\textwidth}
\includegraphics[width=\textwidth]{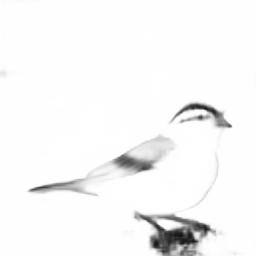}
\end{minipage}
\hfill
\begin{minipage}{0.11\textwidth}
\includegraphics[width=\textwidth]{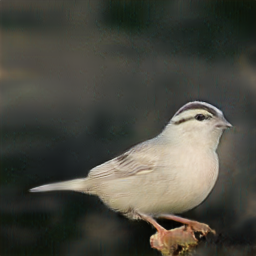}
\end{minipage}
\hfill
\begin{minipage}{0.11\textwidth}
\includegraphics[width=\textwidth]{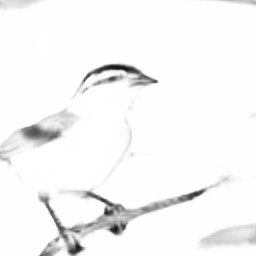}
\end{minipage}
\hfill
\begin{minipage}{0.11\textwidth}
\includegraphics[width=\textwidth]{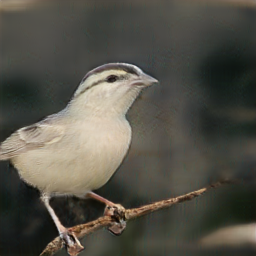}
\end{minipage}
\hfill
\begin{minipage}{0.11\textwidth}
\includegraphics[width=\textwidth]{./fig/fig3/mask_86.png}
\end{minipage}
\hfill
\begin{minipage}{0.11\textwidth}
\includegraphics[width=\textwidth]{./fig/fig3/image_86.png}
\end{minipage}
\hfill
\begin{minipage}{0.11\textwidth}
\includegraphics[width=\textwidth]{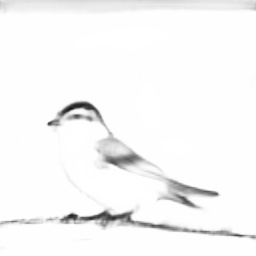}
\end{minipage}
\hfill
\begin{minipage}{0.11\textwidth}
\includegraphics[width=\textwidth]{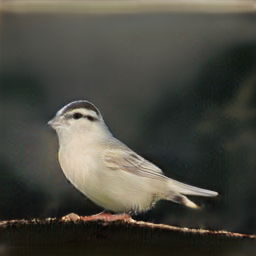}
\end{minipage}
\vfill

\begin{minipage}{0.05\textwidth}
\center{\small{pink}}
\end{minipage}
\hfill
\begin{minipage}{0.11\textwidth}
\includegraphics[width=\textwidth]{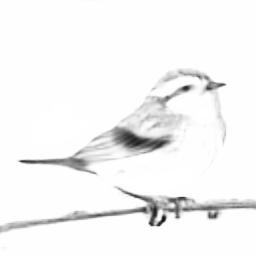}
\end{minipage}
\hfill
\begin{minipage}{0.11\textwidth}
\includegraphics[width=\textwidth]{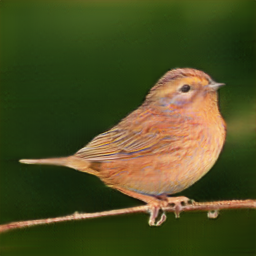}
\end{minipage}
\hfill
\begin{minipage}{0.11\textwidth}
\includegraphics[width=\textwidth]{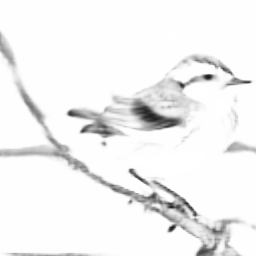}
\end{minipage}
\hfill
\begin{minipage}{0.11\textwidth}
\includegraphics[width=\textwidth]{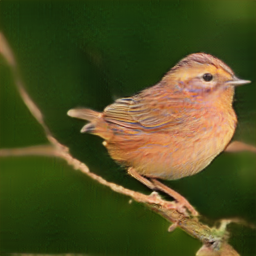}
\end{minipage}
\hfill
\begin{minipage}{0.11\textwidth}
\includegraphics[width=\textwidth]{./fig/fig3/mask_11.png}
\end{minipage}
\hfill
\begin{minipage}{0.11\textwidth}
\includegraphics[width=\textwidth]{./fig/fig3/image_11.png}
\end{minipage}
\hfill
\begin{minipage}{0.11\textwidth}
\includegraphics[width=\textwidth]{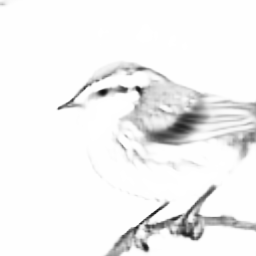}
\end{minipage}
\hfill
\begin{minipage}{0.11\textwidth}
\includegraphics[width=\textwidth]{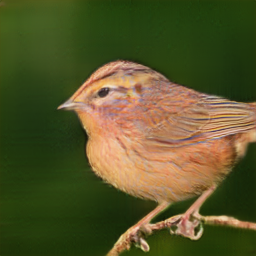}
\end{minipage}
\vfill

\caption{Examples of diverse image generation by changing the color of the input text  from our method on the test set of CUB dataset. The odd columns show the final predicted masks, and the even columns show the corresponding generated images. Best view in color and zoom in.}
\label{fig:cub_change_color}
\vspace{-3mm}
\end{figure*}

\subsection*{Diverse Images Generation from Diverse Texts}
Fig.~\ref{fig:cub_change_color} demonstrates the ability that our method is able to precisely generate images from  diverse text descriptions. For the given text, the color attribute is changed into ``blue'', ``red'', ``white'' and ``pink'' (indicated by the first column), and the corresponding generated images associated with their predicted mask maps (left side) are shown in the same row. The images in the same row are generated by sampling the input noise vectors from the normal distribution. One can observe that the generated images match the given text well (different rows) as well as are diverse from the same text (the same row). It demonstrates that our method is able to generate the images which have the right attributes mentioned in the given text.

In Fig.~\ref{fig:coco_change}, we show that the proposed method is able to generate complex images by modifying the text with respect to objects (first row) or backgrounds (second row). 
From the first row, one can observe that the color of the generated cattle changes into ``brown'' (the 3rd image), and then the cattle are changed into ``sheep'' (last image), corresponding to the modification of the given text.
In the second row, the background ``green grass'' is changed to ``yellow grass'' (the third image) and the ``sky'' is changed to ``sunset'' (last image). All generated images have the corresponding backgrounds.

\begin{figure*}[t!]
\centering

\hspace{-1cm}
\begin{minipage}[c]{0.24\textwidth}
\center{GT}
\end{minipage}
\hfill
\hspace{-1.5cm}
\begin{minipage}[c]{0.24\textwidth}
\center{Input: A herd of black and white cattle standing on a field.}
\end{minipage}
\hfill
\begin{minipage}[c]{0.24\textwidth}
\center{A herd of \textbf{brown} cattle standing on a field.}
\end{minipage}
\hfill
\begin{minipage}[c]{0.24\textwidth}
\center{A herd of black and white \textbf{sheep} standing on a field}
\end{minipage}

\begin{minipage}{0.13\textwidth}
\includegraphics[width=\textwidth]{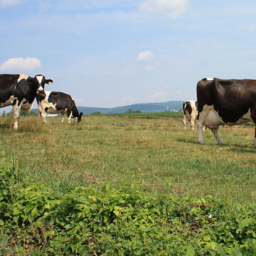}
\end{minipage}
\hfill
\begin{minipage}{0.13\textwidth}
\includegraphics[width=\textwidth]{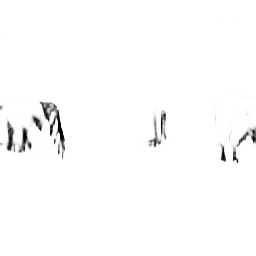}
\end{minipage}
\hfill
\begin{minipage}{0.13\textwidth}
\includegraphics[width=\textwidth]{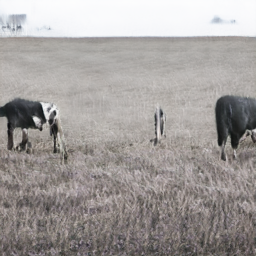}
\end{minipage}
\hfill
\begin{minipage}{0.13\textwidth}
\includegraphics[width=\textwidth]{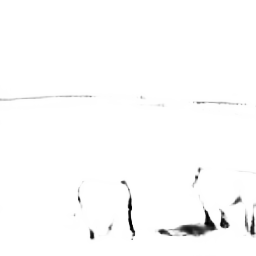}
\end{minipage}
\hfill
\begin{minipage}{0.13\textwidth}
\includegraphics[width=\textwidth]{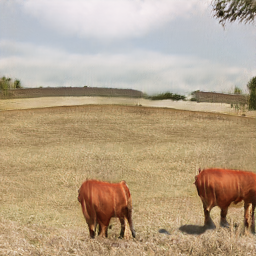}
\end{minipage}
\hfill
\begin{minipage}{0.13\textwidth}
\includegraphics[width=\textwidth]{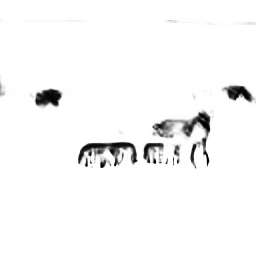}
\end{minipage}
\hfill
\begin{minipage}{0.13\textwidth}
\includegraphics[width=\textwidth]{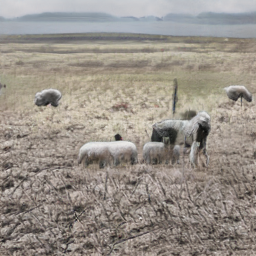}
\end{minipage}
\vspace{8pt}

\hspace{-1cm}
\begin{minipage}[c]{0.24\textwidth}
 \center{GT}
\end{minipage}
\hfill
\hspace{-1.5cm}
\begin{minipage}[c]{0.24\textwidth}
\center{Input: Some horses in a field of green grass with a sky in the background.}
\end{minipage}
\hfill
\begin{minipage}[c]{0.24\textwidth}
\center{Some horses in a field of \textbf{yellow} grass with a sky in the background.}
\end{minipage}
\hfill
\begin{minipage}[c]{0.24\textwidth}
\center{Some horses in a field of green grass with a \textbf{sunset} in the background.}
\end{minipage}

\begin{minipage}{0.13\textwidth}
\includegraphics[width=\textwidth]{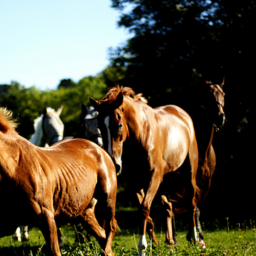}
\end{minipage}
\hfill
\begin{minipage}{0.13\textwidth}
\includegraphics[width=\textwidth]{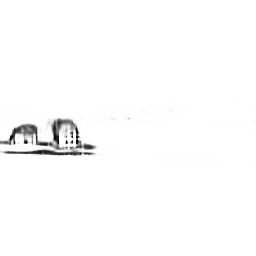}
\end{minipage}
\hfill
\begin{minipage}{0.13\textwidth}
\includegraphics[width=\textwidth]{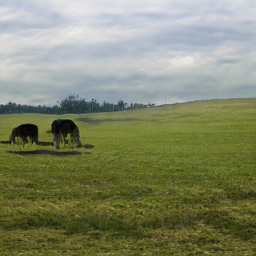}
\end{minipage}
\hfill
\begin{minipage}{0.13\textwidth}
\includegraphics[width=\textwidth]{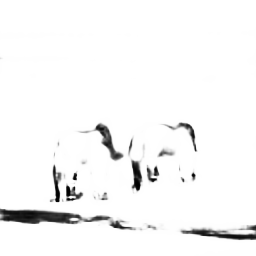}
\end{minipage}
\hfill
\begin{minipage}{0.134\textwidth}
\includegraphics[width=\textwidth]{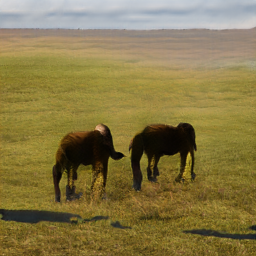}
\end{minipage}
\hfill
\begin{minipage}{0.13\textwidth}
\includegraphics[width=\textwidth]{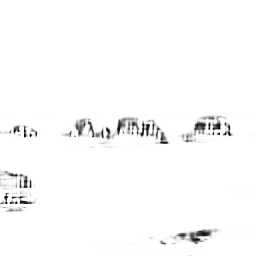}
\end{minipage}
\hfill
\begin{minipage}{0.13\textwidth}
\includegraphics[width=\textwidth]{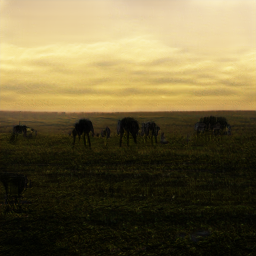}
\end{minipage}
\caption{Example of editing images by changing some words in the input text description (denoted in \textbf{bold} font) on the test set of the COCO dataset. The predicted mask for each generated image is also shown (on its left side). Best view in color and zoom in.}
\label{fig:coco_change}
\end{figure*}

\section*{DAMSM Loss}
\label{sec:damsm}
The DAMSM loss is proposed in \cite{xu2018attngan} which learns the attention model in a semi-supervised way: the supervision is the matching between entire images and whole sentences in both sentence-level and word-level.

Denote that $\bar{e}\in \mathbb{R}^{256}$ is the sentence feature vector, and $e\in \mathbb{R}^{256\times18}$ is a matrix in which the $i$-th column $e_i$ of $e$ is the feature vector of the $i$-th word.
An Inception-v3 model \cite{szegedy2016rethinking} pretrained on ImageNet \cite{russakovsky2015imagenet} is used as image encoder that extract the local feature matrix $f\in \mathbb{R}^{768\times289}$ by reshaping the feature maps in shape of $768\time17\time17$ output from the ``mixed 6e'' layer. Each column of $f$ represents a sub-region of the image.
Meanwhile, the global feature vector $\bar{f}\in \mathbb{R}^{2048}$ is extracted from the last average pooling layer of Inception-v3. Finally, the image features are converted to a common semantic space of text features by adding a perceptron layer:
\begin{equation}
\begin{split}
    v &= Wf, v\in \mathbb{R}^{D\times289}\\
    \bar{v} &= \bar{W}\bar{f}, \bar{v}\in \mathbb{R}^{D}
\end{split}
\end{equation}
where $D$ is the dimension of the image and text feature space. The $i$-th column $v_i$ is the visual feature vector for the $i$-th sub-region of the image.

The similarity matrix for all possible pairs of words in the sentence and sub-regions in the image are calculated by:
\begin{equation}
    s = e^Tv, s\in \mathbb{R}^{T\times289}
\end{equation}
where $s_{i,j}$ is the dot-product similarity between the $i$-th word of the sentence and the $j$-th sub-region of the image. Then, the similarity matrix is normalized as follows:
\begin{equation}
    \bar{s}_{i,j} = \frac{exp(s_{i,j})}{\Sigma^{T-1}_{k=0}exp(s_{k,j})}
\end{equation}
 
Then, an attention model is built to compute a region context vector for each word (query). The region-context vector $c_i$ is a dynamic representation of the image’s subregions related to the $i$-th word of the sentence. It is computed as the weighted sum over all regional visual vectors:
\begin{equation}
    c_i = \sum^288_{j=0}\alpha_i v_j, where \alpha_i=\frac{exp(\gamma_1\bar{s}_{i,j})}{\Sigma^{288}_{k=0}exp(\gamma1\bar{s}_{i,k})}
\end{equation}
Here $\gamma_1=5$ is a factor that determines how much attention is paid to features of its relevant sub-regions when computing the region-context vector for a word.

Finally, the relevance between the $i$-th word and the image using the cosine similarity between $c_i$ and $e_i$ is defined as:
\begin{equation}
    R(c_i,e_i) = (c^T_ie_i)/(\left\|c_i\right\|\left\|e_i\right\|)
\end{equation}
The attention-driven image-text matching score between the entire image (Q) and the whole text
description (D) is defined as:
\begin{equation}
    R(Q,D) = log \left( \sum^{T-1}_{i=1}exp(\gamma_2R(c_i,e_i))\right)^{\frac{1}{\gamma_2}}\label{eq:r}
\end{equation}
where $\gamma_2=5$ is a factor that determines how much to magnify the importance of the most relevant word-to-region context pair.

For a batch of image-sentence pairs ${(Q_i,D_i)}^M_{i=1}$, $M=10$, the posterior probability of sentence $D_i$ being matching with image $Q_i$ is computed as:
    $P(D_i|Q_i) = \frac{exp(\gamma_3R(Q_i,D_i))}{\Sigma^{M}_{j=1}exp(\gamma3(Q_i,D_j))},$
where $\gamma_3=10$ is a smoothing factor determined by experiments. In this batch of sentences, only $D_i$ matches the image $Q_i$, and treat all other mismatching sentences as mismatching descriptions. The loss function is defined as the negative log posterior probability that the images are matched with their corresponding text descriptions (ground truth):
\vspace{-3mm}
\begin{align}
     \mathcal{L}^w_1 &= -\sum^M_{i=1}log P(D_i|Q_i),\\
     \mathcal{L}^w_2 &= -\sum^M_{i=1}log P(Q_i|D_i),
\end{align}
where `w' stands for ``wor''. $\mathcal{L}^w_1$ and $\mathcal{L}^w_2$ are symmetrical. $ P(Q_i|D_i)$ is the posterior probability that sentence $D_i$ is matched with its corresponding image $Q_i$.
To compute the loss between sentence vector $\bar{e}$ and global image vector $\bar{v}$, $\mathcal{L}^s_1$ and $\mathcal{L}^s_2$, Eq.~\eqref{eq:r} as $R(Q,D) = (\bar{v}^T\bar{e}_i)/(\left\|\bar{v}\right\|\left\|\bar{e}\right\|)$.

Finally, the DAMSM loss is defined as:
\vspace{-3mm}
\begin{equation}
     \mathcal{L}_{DAMSM} = \mathcal{L}^w_1+\mathcal{L}^w_2+\mathcal{L}^s_1+\mathcal{L}^s_2.
\end{equation}

{\small
\bibliographystyle{ieee_fullname}
\bibliography{egbib}
}
\end{document}